\documentclass[10pt,journal,compsoc]{IEEEtran}
\ifCLASSOPTIONcompsoc
  \usepackage[nocompress]{cite}
\else
  \usepackage{cite}
\fi
\ifCLASSINFOpdf
\else
\fi
\usepackage{booktabs}
\usepackage{subcaption}
\usepackage{multirow}
\usepackage{mmstyle}
\usepackage{mathtools}
\usepackage{hanging}
\usepackage{tablefootnote}
\usepackage{epsfig}
\usepackage{graphicx}
\usepackage{bmpsize}
\usepackage{amsmath}
\usepackage{amssymb}
\usepackage{url}
\newcommand{\algname}{CARAFE++}

\begin{document}
\title{CARAFE++: Unified Content-Aware ReAssembly of FEatures}

\author{Jiaqi~Wang,~Kai~Chen,~Rui~Xu,~Ziwei~Liu,~Chen~Change~Loy,~and~Dahua~Lin% <-this % stops a space
  \IEEEcompsocitemizethanks{
    \IEEEcompsocthanksitem Jiaqi~Wang is with The Chinese University of Hong Kong, Hong Kong. E-mail: wj017@ie.cuhk.edu.hk.
    \IEEEcompsocthanksitem Chen~Kai is with The Chinese University of Hong Kong, Hong Kong. E-mail: ck015@ie.cuhk.edu.hk.
    \IEEEcompsocthanksitem Rui~Xu is with The Chinese University of Hong Kong, Hong Kong. E-mail: xr018@ie.cuhk.edu.hk.
    \IEEEcompsocthanksitem Ziwei~Liu is with Nanyang Technological University, Singapore. E-mail: ziwei.liu@ntu.edu.sg.
    \IEEEcompsocthanksitem Chen~Change~Loy is with Nanyang Technological University, Singapore. E-mail: ccloy@ntu.edu.sg.
    \IEEEcompsocthanksitem Dahua~Lin is with The Chinese University of Hong Kong, Hong Kong. E-mail: dhlin@ie.cuhk.edu.hk.}% <-this % stops an unwanted space
}

\IEEEtitleabstractindextext{%
  % !TEX root = ../main.tex

\begin{abstract}

    Feature reassembly, \ie feature downsampling and upsampling, is a key operation in a number of modern convolutional network architectures, \eg,~residual networks and feature pyramids.
    Its design is critical for dense prediction tasks such as object detection and semantic/instance segmentation.
    In this work, we propose unified Content-Aware ReAssembly of FEatures (CARAFE++), a universal, lightweight and highly effective operator to fulfill this goal.
    CARAFE++ has several appealing properties:
    (1) Unlike conventional methods such as pooling and interpolation that only exploit sub-pixel neighborhood, CARAFE++ aggregates contextual information within a large receptive field.
    (2) Instead of using a fixed kernel for all samples (\eg convolution and deconvolution), CARAFE++ generates adaptive kernels on-the-fly to enable instance-specific content-aware handling.
    (3) CARAFE++ introduces little computational overhead and can be readily integrated into modern network architectures.
    We conduct comprehensive evaluations on standard benchmarks in object detection, instance/semantic segmentation and image inpainting.
    CARAFE++ shows consistent and substantial gains across all the tasks (2.5\% $AP_{box}$, 2.1\% $AP_{mask}$, 1.94\% mIoU, 1.35 dB respectively) with negligible computational overhead.
    It shows great potential to serve as a strong building block for modern deep networks.

\end{abstract}

  }

\maketitle

\IEEEdisplaynontitleabstractindextext

% \fi
\IEEEpeerreviewmaketitle

% !TEX root = ../main.tex

\IEEEraisesectionheading{\section{Introduction}
	\label{sec:intro}}

\IEEEPARstart{F}{eature} reassembly, i.e., downsampling and upsampling, is one of the most fundamental operations in deep neural networks.
On the one hand, in dense prediction tasks (\eg, super resolution~\cite{Dong_2016, Lim_2017}, inpainting~\cite{iizuka2017globally, pathak2016context} and semantic segmentation~\cite{zhao2017pyramid, Chen_2018}), the input image is downsampled in the encoders to enlarge receptive field, gather semantic information, and reduce computational cost. For the decoders, the high-level/low-resolution feature map is upsampled to match the high-resolution supervision.
On the other hand, feature reassembly is also involved in fusing a high-level/low-resolution feature map with a low-level/high-resolution feature map, which is widely adopted in many state-of-the-art architectures, \eg, Feature Pyramid Network~\cite{lin2017feature}, U-Net~\cite{ronneberger2015u} and Stacked Hourglass~\cite{Newell_2016}.
Therefore, designing an effective feature reassembly operator becomes a critical issue.

\begin{figure}
	\centering
	\includegraphics[width=0.95\linewidth]{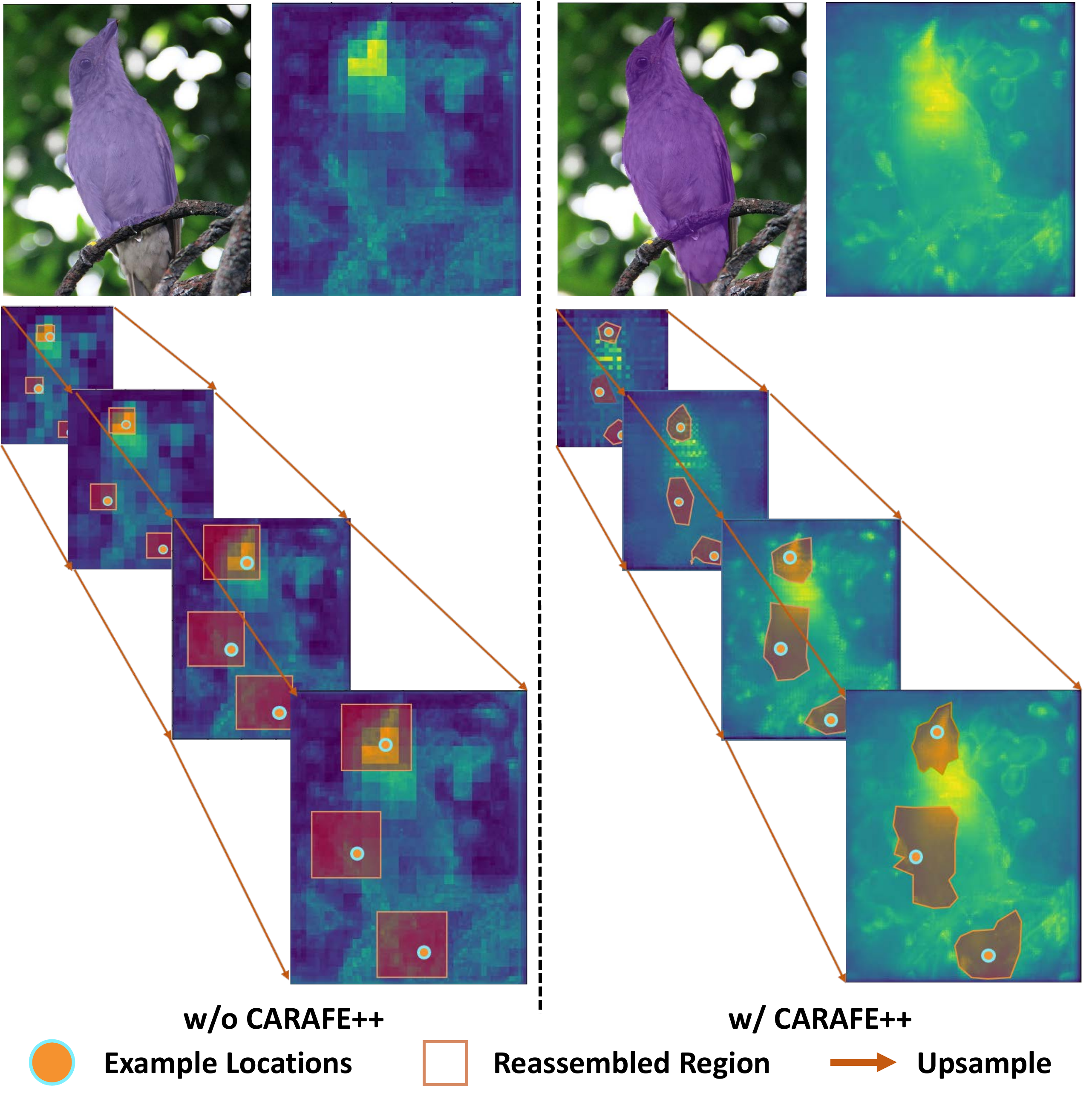}
	\vspace{-10pt}
	\caption{\small{\textbf{Illustration of \algname~working mechanism}. \textbf{Left:} Multi-level FPN features from Mask R-CNN baseline (left to dotted line) and \textbf{Right:} Multi-level FPN features from Mask R-CNN with \algname (right to dotted line).
			For sampled locations, this figure shows the accumulated reassembled regions in the top-down pathway of FPN.
			Information inside such a region is reassembled into the corresponding reassembly center.
		}
	}
	\label{fig:features}
	\vspace{-20pt}
\end{figure}

Feature reassembly is composed of downsampling and upsampling on feature maps. Pooling and interpolation are the most widely adopted operator families for downsampling and upsampling, respectively. The max pooling and average pooling are two representative pooling methods, which reassemble features inside a local region with a hand-crafted kernel.
The nearest neighbor and bilinear interpolations are the most commonly used upsampling operators, which adopt spatial distance between pixels to guide the upsampling process.
However, both pooling and interpolation are rule-based operators, which fail to capture the rich semantic information required by dense prediction tasks.

An alternative method for adaptive feature reassembly is convolution and deconvolution~\cite{Noh_2015}. A convolution layer adjusts `stride' to control the spatial distance to apply convolution kernel. A deconvolution layer works as an inverse operator of a convolution layer, which learns a set of instance-agnostic upsampling kernels.
However, there exist two major drawbacks.
First, a convolution/deconvolution operator applies the same kernel across the entire image, regardless of the underlying content. This restricts its capability in responding to local variations. Second, it comes with heavy computational workload when a large kernel size is used. This makes it difficult to cover a larger region that goes beyond a small neighborhood, thus limiting its expressive power and performance.

In this work, we move beyond these limitations, and seek a feature reassembly operator that is capable of 1) aggregating information within large receptive field, 2) adapting to instance-specific contents on-the-fly, and 3) maintaining computation efficiency.
To this end, we propose a lightweight yet highly effective operator, called \emph{unified Content-Aware ReAssembly of Features (CARAFE++)}.
Specifically, \algname~reassembles the features inside a predefined region via a weighted combination, where the weights are generated in a content-aware manner.
A reassembly kernel is generated for each \emph{target} location. And the feature reassembly is performed on the corresponding location of the \emph{input} feature map.

Note that these spatially adaptive weights are not learned as network parameters.
Instead, they are predicted on-the-fly, using a lightweight fully-convolutional module with softmax activation.
Figure~\ref{fig:features} reveals the working mechanism of \algname~in an upsampling case.
We visualize the feature maps in the top-down pathway of feature pyramid network (FPN)\cite{lin2017feature} and compare \algname
with the nearest neighbor interpolation baseline.
After upsampled by \algname, a feature map can represent the shape of an object more accurately, so that the model can predict better instance segmentation results. Our \algname~not only rescales the feature map spatially, but also learns to enhance its discrimination.

To demonstrate the universal effectiveness of \algname, we conduct comprehensive evaluations across a wide range of dense prediction tasks, \ie, object detection, instance segmentation, semantic segmentation, image inpainting, with mainstream architectures.
With negligible extra cost, \algname~boosts the performance of Faster R-CNN~\cite{ren2015faster} by 2.5\% $AP_{box}$ in object detection and Mask R-CNN~\cite{he2017mask} by 2.1\% $AP_{mask}$ in instance segmentation on MS COCO~\cite{lin2014microsoft} test-dev.
\algname~further improves UperNet~\cite{xiao2018unified} by 1.94\% mIoU on ADE20k~\cite{zhou2017scene, zhou2018semantic} val in semantic segmentation, and improves Global\&Local~\cite{iizuka2017globally} by 1.35 dB of PSNR on Places~\cite{zhou2017places} val in image inpainting.
The substantial gains on all the tasks demonstrate that \algname~is an effective and efficient feature reassembly operator that has great potential to serve as a strong building block for future research.

In a previous conference version in ICCV 2019, we proposed CARAFE~\cite{Wang_2019_ICCV}, which is an effective and efficient upsampling operator. \algname~shares a similar design while it is more universal. \algname~can be readily integrated to networks for both upsampling and downsampling.
We perform an extensive evaluation of \algname~when it is applied in a wide range of network architectures.
Experimental results show that adopting \algname~in both upsampling and downsampling can consistently and substantially outperforms CARAFE\footnote{Adopting CARAFE is equivalent to adopting \algname~for upsampling.} on object detection, instance segmentation, semantic segmentation and image inpainting.


% !TEX root = ../main.tex

\section{Related Work}
\label{sec:related}

\noindent
\textbf{Downsampling \& Upsampling Operators.}
Downsampling \& upsampling operators are basic building blocks of convolutional neural network architectures.
Contemporary convolution networks usually downsample the input features in the first few layers.
Feature upsampling operators are essential for convolutional networks to make high-quality dense predictions.

Among various downsampling operators, max pooling and average pooling are the most widely used choices. They reassemble features in a rule-based manner. For upsampling, nearest neighbor and bilinear interpolations are representative methods.
These interpolations leverage distances to measure the correlations between pixels, and hand-crafted upsampling kernels are used in them.
Several methods are proposed to downsample and upsample a feature map using learnable operators.
For example, convolution and deconvolution (transposed convolution)~\cite{Noh_2015} are the most representative ones among those learnable upsamplers.
Pixel Shuffle~\cite{Shi_2016} proposes an upsampling operator that reshapes depth of the channel space into width and height of the spatial space.
Guided Upsampling (GUM)\cite{mazzini2018guided} performs interpolation by sampling pixels with learnable offsets.
In the downsampling pipeline, detail-preserving pooling (DPP)\cite{Saeedan_2018_CVPR}~is proposed to preserve more detail information by focusing on local spatial changes of pixels in a sliding window.
Local Importance-based Pooling (LIP)\cite{Gao_2019_ICCV}~is a recently proposed downsampling operator, which performs a weighted average pooling.
However, these methods either exploit contextual information in a small neighborhood, or require expensive computation to perform adaptive downsampling and upsampling.
Within the realms of super-resolution and denoising, some other works~\cite{mildenhall2018burst,jo2018deep,hu2019meta} also explore the use of learnable kernels spatially in low-level vision.
In this study, we demonstrate the effectiveness and working mechanism of unified content-aware feature reassembly for downsampling and upsampling in several visual perception tasks, and provide a lightweight solution.

\begin{figure*}[t]
    \centering
    \includegraphics[width=0.9\linewidth]{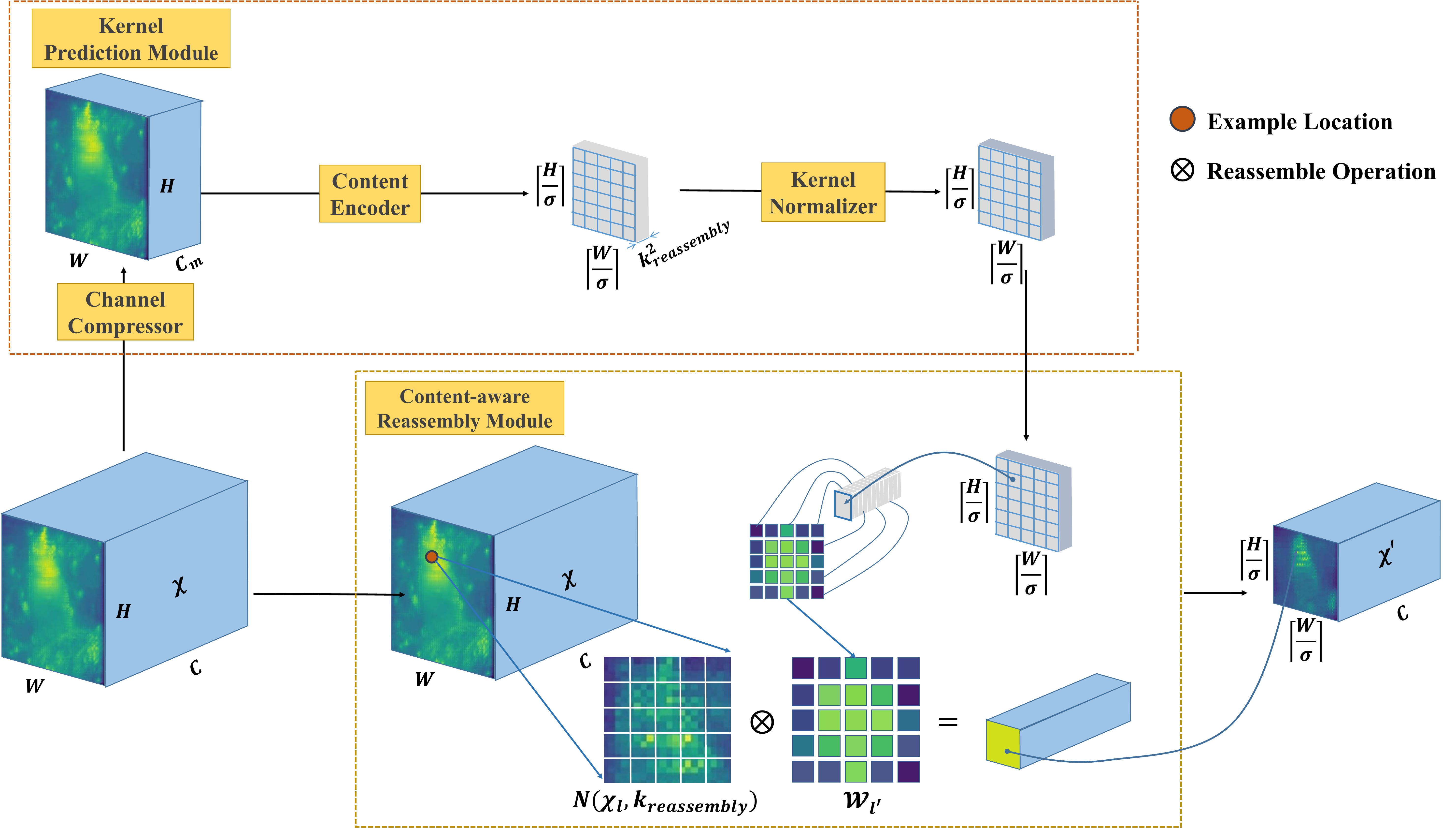}
    \caption{\small{\textbf{The overall framework of {\algname} for downsampling}. {\algname} is composed of two key components, \ie, kernel prediction module and content-aware reassembly module. A feature map with size $C \times H\times W$ is downsampled by a factor of $\sigma(=2)$ in this figure.}}
    \label{fig:framework_down}
    \vspace{-10pt}
\end{figure*}

\noindent
\textbf{Dense Prediction Tasks.}
Object detection~\cite{ren2015faster,lin2017focal,liu2016ssd,wang2019region,SABL_2020_ECCV} is the task of localizing objects with bounding-boxes, instance segmentation~\cite{he2017mask} further requires the prediction of instance-wise masks.
Many studies~\cite{lin2017feature,liu2018path,kong2018deep,zhao2019m2det} exploit multi-scale feature
pyramids to deal with objects at different scales.
By adding extra mask prediction branches, Mask R-CNN~\cite{he2017mask} and
its variants~\cite{chen2019hybrid} yield promising pixel-level results.
Semantic segmentation~\cite{liu2015semantic,li2017not} requires pixel-wise semantic prediction for given images.
PSPNet~\cite{zhao2017pyramid} introduces spatial pooling at multiple grid scales.
UperNet~\cite{xiao2018unified} designs a more generalized framework based on PSPNet.
These detection and segmentation methods usually adopt backbones, \eg, ResNet~\cite{he2016deep}, which are pretrained on image classification dataset. The input features are downsampled for several times in such image classification networks.
Image or video inpainting~\cite{yu2018generative,Xu_2019_CVPR,ulyanov2018deep} is a classical problem that aims at completing missing regions of images.
%,
U-net~\cite{ronneberger2015u}, which adopts multiple downsampling and upsampling operators, is popular among recent works~\cite{iizuka2017globally,ulyanov2018deep}.
Liu \etal.~\cite{liu2018image} introduce partial convolution layer to alleviate the influence of missing regions on the convolution layers.
By replacing the downsampling and upsampling operators in the aforementioned networks, our \algname~demonstrates consistent effectiveness across a wide range of dense prediction tasks.

% !TEX root = ../main.tex
\begin{figure*}[t]
    \centering
    \includegraphics[width=0.9\linewidth]{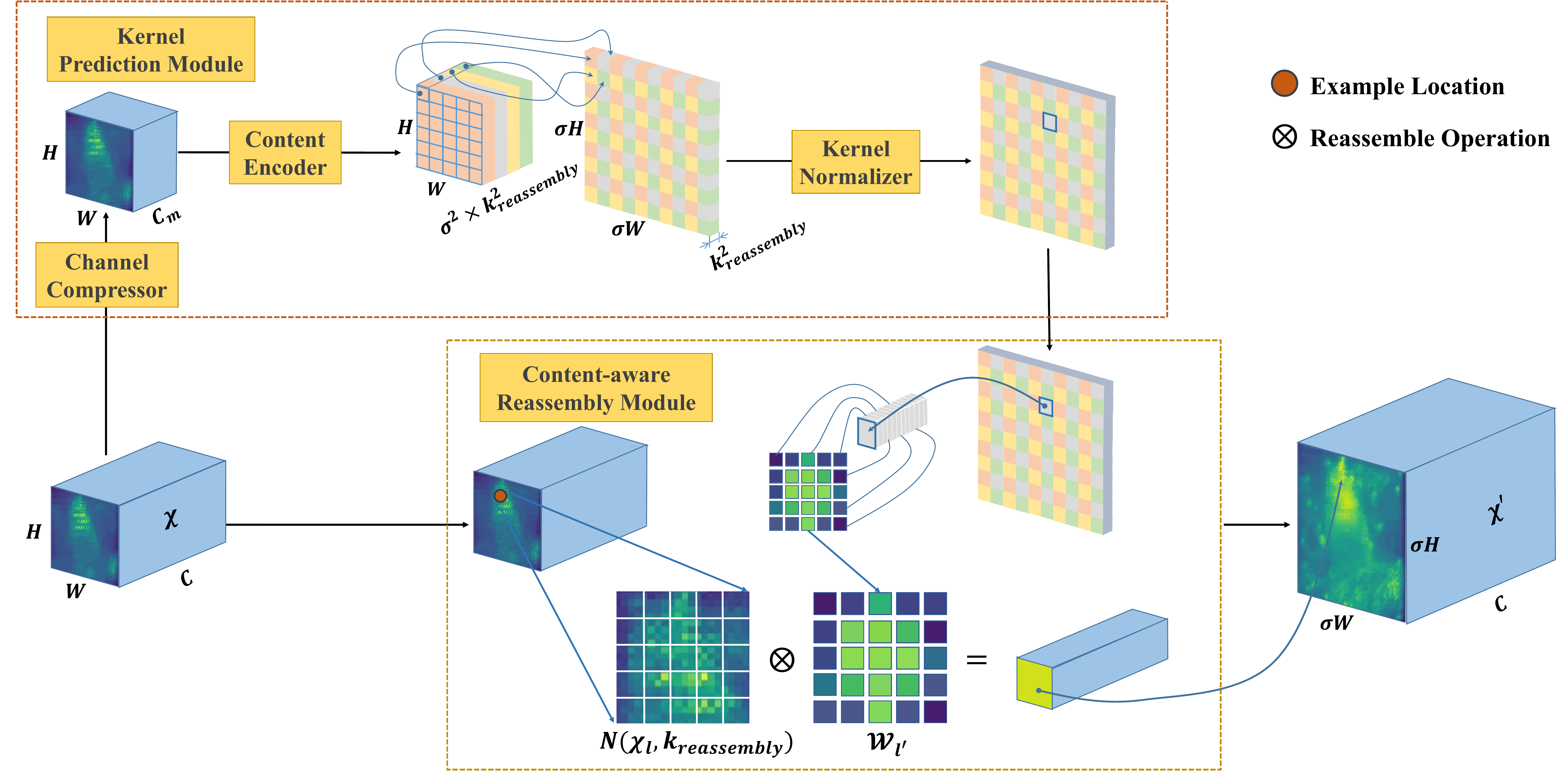}
    \caption{\small{\textbf{The overall framework of {\algname} for upsampling}. A feature map with size $C \times H\times W$ is upsampled by a factor of $\sigma(=2)$ in this figure.}}
    \label{fig:framework}
    \vspace{-10pt}
\end{figure*}

\section{Content-Aware ReAssembly of FEatures}
\label{sec:methodology}

Feature reassembly, \ie, downsampling, upsampling, is a key operator in many modern convolutional network architectures developed for tasks including object detection, instance segmentation, and scene parsing.
In this work, we propose Content-Aware ReAssembly of FEatures (\algname) to reassemble a feature map. On each location, \algname~leverages the underlying content information to predict reassembly kernels and reassemble the features inside a predefined nearby region.
Thanks to this unique capability, \algname~achieves better performance than the mainstream downsampling and upsampling operators, \eg,~pooling or interpolation.

\subsection{Formulation}
\label{subsec:formulation}
\algname~works as a reassembly operator with content-aware kernels.
It consists of two steps. The first step is to predict a reassembly kernel for
each target location according to its content, and the second step is to reassemble
the features with the predicted kernels.

Given a feature map $\cX$ of size $C \times H\times W$ and a resizing ratio $\sigma$
(supposing $\sigma$ is an integer), \algname~will produce a new feature map
$\cX^\prime$ of size $C \times \lceil H /\sigma \rceil \times\lceil W /\sigma \rceil $ in the downsampling process and $C \times \sigma H\times\sigma W$ in the upsampling process.
For any target location $l^\prime=(i^\prime,j^\prime)$ of the output
$\cX^\prime$, there is a corresponding source location $l=(i,j)$ at the input $\cX$,
where $i=\sigma i^\prime,j=\sigma j^\prime$ for downsampling, and $i=\left \lfloor i^\prime/\sigma \right \rfloor,j=\left \lfloor j^\prime/\sigma \right \rfloor$ for upsampling.
Here we denote $N(\cX_l, k)$ as the $k\times k$ sub-region of $\cX$ centered at the location $l$,
\ie, the neighbor of $\cX_l$.

In the first step, the \emph{kernel prediction module} $\psi$ predicts a
location-wise kernel $\cW_{l^\prime}$ for each location $l^\prime$,
based on the neighbor of $\cX_l$, as shown in Eqn.~\eqref{eq:kernel-prediction}.
The reassembly step is formulated as Eqn.~\eqref{eq:reassembly}, where $\phi$
is the \emph{content-aware reassembly module} that reassembles the neighbor of $\cX_l$
with the kernel $\cW_{l^\prime}$:
\begin{equation} \label{eq:kernel-prediction}
    \cW_{l^\prime} = \psi(N(\cX_l, k_{encoder})).
\end{equation}
\begin{equation} \label{eq:reassembly}
    \cX^\prime_{l^\prime} = \phi(N(\cX_l, k_{reassembly}), \cW_{l^\prime}).
\end{equation}
We specify the details of $\psi$ and $\phi$ in the following parts.

\subsection{Kernel Prediction Module}
\label{subsec:kernel prediction module}

The kernel prediction module is responsible for generating the reassembly kernels in a content-aware manner.
Each target location corresponds to a source location and requires a $k_{reassembly}\times k_{reassembly}$ reassembly kernel, where $k_{reassembly}$ is the reassembly kernel size.
Therefore, this module will output the reassembly kernels of size $C_{reassembly} \times \lceil H /\sigma \rceil \times\lceil W /\sigma \rceil$  for downsampling, and $C_{reassembly} \times \sigma H \times \sigma W$ for upsampling,
where $C_{reassembly}=k_{reassembly}^2$.

The kernel prediction module is composed of three submodules, \ie,
\emph{channel compressor}, \emph{content encoder} and \emph{kernel normalizer},
as shown in Figure~\ref{fig:framework_down} and Figure~\ref{fig:framework}.
The channel compressor reduces the channel of the input feature map.
The content encoder then takes the compressed feature map as input and encodes
the content to generate reassembly kernels.
Lastly, the kernel normalizer applies a softmax function to each reassembly kernel.
The three submodules are explained in detail as follows.

\noindent\textbf{Channel Compressor}.
We adopt a $1\times 1$ convolution layer to compress the input feature channel from $C$ to $C_m$.
Specifically, we adopt $C_m=16$ for downsampling and $C_m=64$ for upsampling in experiments.
Reducing the channel of input feature map leads to less computational cost in the following steps, making \algname~much more efficient.
It is also possible to use larger kernel sizes for the content encoder under the same budget.
Experimental results show that reducing the feature channel in an acceptable
range will not harm the performance.

\noindent\textbf{Content Encoder}.
We use a convolution layer of kernel size $k_{encoder}$ to generate reassembly
kernels based on the content of input features. The input channel of this convolution layer is $C_m$. Certain settings are different for downsampling and upsampling.

For downsampling, the stride of this convolution layer is $\sigma$.
Thus, the predicted reassembly kernel has a size of $C_{reassembly} \times \lceil H /\sigma \rceil \times\lceil W /\sigma \rceil$.
For upsampling, the output channels of this convolution layer is $\sigma^2C_{reassembly}$. It is then reorganized from depth of the channel space into width and height of the spatial space. Finally, the predicted reassembly kernel has a size of $C_{reassembly} \times \sigma H \times \sigma W$.

Intuitively, increasing $k_{encoder}$ enlarges the receptive field of the
encoder, and this allows one to exploit contextual information within a larger region,
which is important for predicting the reassembly kernels.
However, the computational complexity grows quadratically with the kernel size,
while the benefits from a larger kernel size may not match equally.
An empirical formula, $k_{encoder} = k_{reassembly}-2$, is a good trade-off between
performance and efficiency through our study in Section~\ref{subsec:ablation}.

\noindent\textbf{Kernel Normalizer}.
Before being applied to the input feature map, each
reassembly kernel with size of $k_{reassembly}\times k_{reassembly}$ is normalized with a softmax function spatially.
The normalization step forces the sum of kernel values to 1. The normalized values result a soft selection across a local region.
Due to the kernel normalizer, \algname~does not perform any rescaling and change to
the mean values of the feature map, but reassembles features spatiallly.

\subsection{Content-aware Reassembly Module}
\label{subsec:content-aware reassembly module}

With each reassembly kernel $\cW_{l^\prime}$, the content-aware reassembly module
reassembles the features within a local region via the function $\phi$.
We adopt a simple form of $\phi$, which is just a weighted sum operator.
For a target location $l^\prime$ and the corresponding square region $N(\cX_l, k_{reassembly})$ centered at $l=(i,j)$, the reassembly
is shown in Eqn.~\eqref{eq:equal_3}, where $r = \left \lfloor k_{reassembly} / 2 \right \rfloor$:
\begin{equation} \label{eq:equal_3}
    \begin{aligned}
        \cX^\prime_{l^\prime}=
        \sum_{n = -r}^{r}\sum_{m = -r}^{r} \cW_{l^\prime(n,m)} \cdot \cX_{(i+n, j+m)}.
    \end{aligned}
\end{equation}

With the reassembly kernel, each pixel in the region of $N(\cX_l, k_{reassembly})$
contributes to the target pixel $l^\prime$ differently, based on the content
of features.
The semantics of the reassembled feature map can be stronger than the original
one, since the information from relevant points in a local region can be more attended.

\subsection{Relation to Previous Operators}
Here we discuss the relations between \algname~and dynamic filter~\cite{brab2016dynamic}, spatial attention~\cite{chen2017sca}, local importance-based pooling~\cite{Gao_2019_ICCV}, spatial transformer~\cite{jaderberg2015spatial} and deformable convolution~\cite{Dai_2017}, which share similar design philosophy but with different focuses.

\noindent
\textbf{Dynamic Filter.}
Dynamic filter generates instance-specific convolutional filters conditioned on the input of the network, and then applies the predicted filter on the input.
Both dynamic filter and \algname~are content-aware operators, but a fundamental difference between them lies at their kernel generation process.
Specifically, dynamic filter works as a two-step convolution, where the additional dynamic filter prediction step requires heavy computation.
On the contrary, \algname~is simply a reassembly of features in local regions, without learning the feature transformation across channels.
Supposing the channels of input feature map is $C$ and kernel size of the
filter is $K$, then the predicted kernel weights for each location is
$C\times C\times K\times K$ in dynamic filter. For \algname, the kernel weights is only $K\times K$. Thus, it is more efficient in memory and speed.

\noindent
\textbf{Spatial Attention.}
Spatial attention predicts an attention map with the same spatial size as the input feature, and then rescales the feature map on each location.
Our \algname~reassembles the features in a local region by weighted summation.
In summary, spatial attention is a rescaling operator with point-wise guidance while \algname~is a reassembly operator with region-wise local guidance.
Spatial attention can be seen as a special case of \algname~where the reassembly kernel size is 1, regardless of the kernel normalizer.

\noindent
\textbf{Local Importance-Based Pooling (LIP).}
Local Importance-Based Pooling (LIP) is a recently proposed pooling method, which combines the self-attention and average pooling. In LIP, an attention map with the same shape (\ie, spatial size and channels) as the input feature map is predicted by a series of convolution layers. The predicted attention map is normalized by a sigmoid function, and then directly multiplied by the original feature map. A standard average pooling downsamples the feature map after self-attention. The self-attention in LIP is a combination of spatial attention and channel attention. Similar to spatial attention, it fails to adaptively reassemble information from a large receptive field. To predict the attention map with the same shape as the input, LIP requires a series of convolution layers with deep output channels and brings heavy computational cost. Moreover, LIP is not available for upsampling.

\noindent
\textbf{Spatial Transformer Networks (STN).}
STN predicts a global parametric transformation conditioned on the input feature map and warps the feature via the transformation.
However, this global parametric transformation assumption is too strong to represent complex spatial variance; and STN is known to be instable to train.
Here, \algname~uses the location-specific reassembly to handle the spatial relations, which enables more flexible local geometry modeling.

\noindent
\textbf{Deformable Convolutional Networks (DCN).}
DCN also adopts the idea of learning geometric transformation and combines it with the regular convolution layers.
It predicts kernel offsets other than using grid convolution kernels.
Similar to dynamic filter, it is also a heavy parametric operator with much more computational cost than \algname.

% !TEX root = ../main.tex

\section{Applications of \algname}
\label{sec:applications}

\algname~can be seamlessly integrated into existing frameworks where downsampling \& upsampling operators are needed.
Here we present some applications in mainstream dense prediction tasks. With negligible additional computational cost, \algname~benefits mainstream methods in
both high-level and low-level tasks, such as object detection, instance segmentation, semantic segmentation and image inpainting.

\begin{figure}
	\centering
	\includegraphics[width=\linewidth]{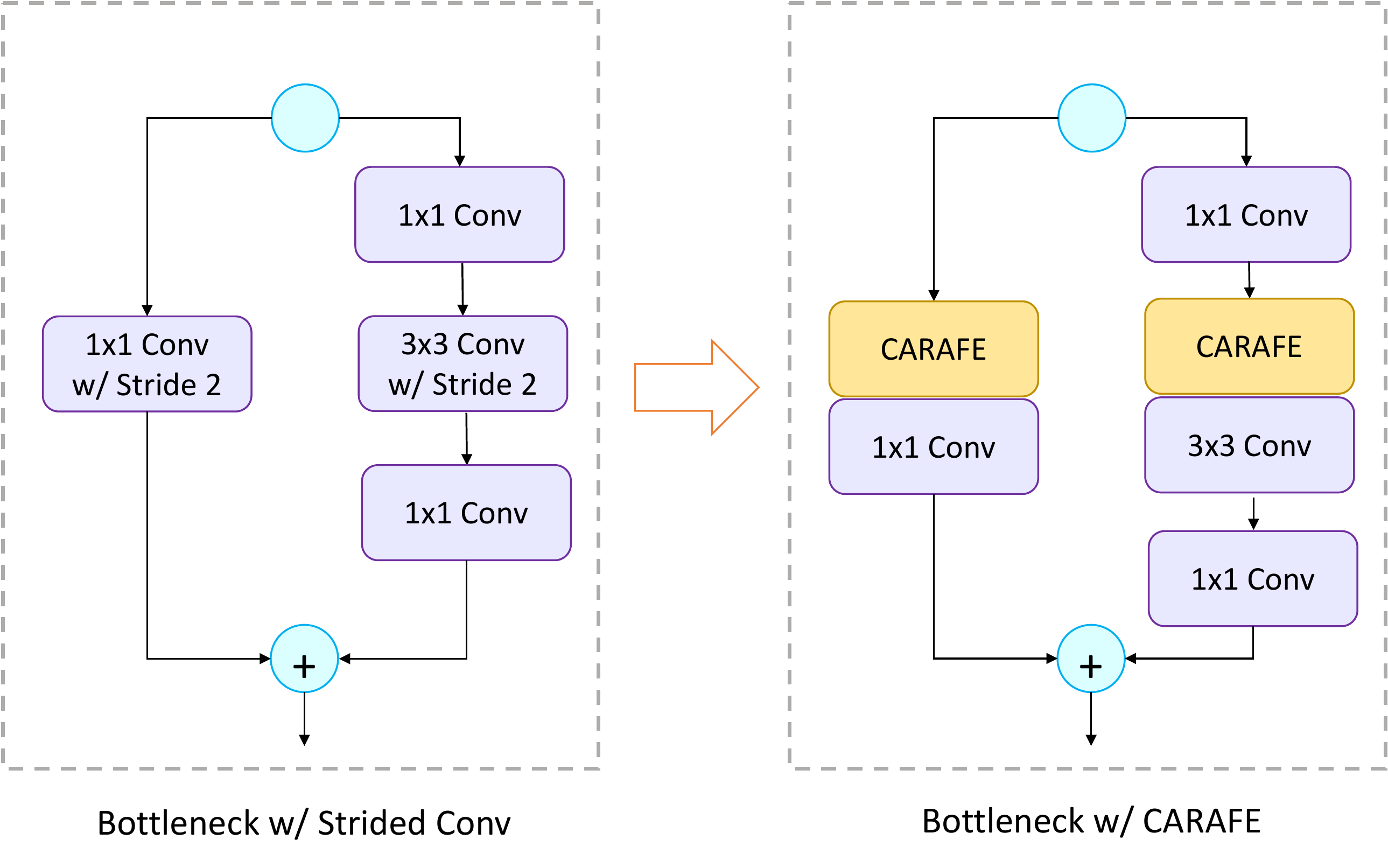}
	\caption{\small{\textbf{Bottleneck architecture with \algname~in ResNet}. \algname~can be readily integrated into Bottleneck with strided convolution. \algname~downsamples the input feature map by 2x. And convolution layers with the same kernel size but no stride are applied afterwards.}}
	\label{fig:bottleneck}
	\vspace{-10pt}
\end{figure}

\subsection{Object Detection and Instance Segmentation}
\label{subsec:application in object detection and instance segmentation}

\noindent\textbf{ResNet Backbone.} Backbone is the cornerstone of object detection. The backbone with strided convolution layers is adopted to downsample input images and produce pyramid features.
The backbone applied in object detection and instance segmentation is generally pretrained on image classification dataset, \eg, ImageNet.

ResNet backbone~\cite{he2016deep}~is one of the most widely used backbones in object detection. ResNet has five blocks, named as Res-1, Res-2, Res-3, Res-4 and Res-5, respectively. Res-1 contains one 3x3 convolution layer and one max pooling layer, it is also called as the stem layer. Res-2, Res-3, Res-4 and Res-5 are stacked building blocks. A feature map is downsampled in Res-3, Res-4 and Res-5 at their first building block with strided convolution layers. CARAFE++ can be readily integrated into ResNet. As shown in Figure~\ref{fig:bottleneck},~\algname~is applied to the building block (named as Bottleneck) with strided convolution layers. In the original design of bottleneck, convolution layers with stride downsamples the feature map by 2x directly. In the bottleneck w/~\algname, \algname~downsamples the feature map before convolution layers.
Furthermore, \algname~replaces the max pooling which downsamples the feature map by 2x in Res-1 as well. In summary, there are seven~\algname~are applied in a ResNet backbone, \ie, one in Res-1, two in each of Res-3, Res-4 and Res-5.

\begin{figure}
	\centering
	\includegraphics[width=\linewidth]{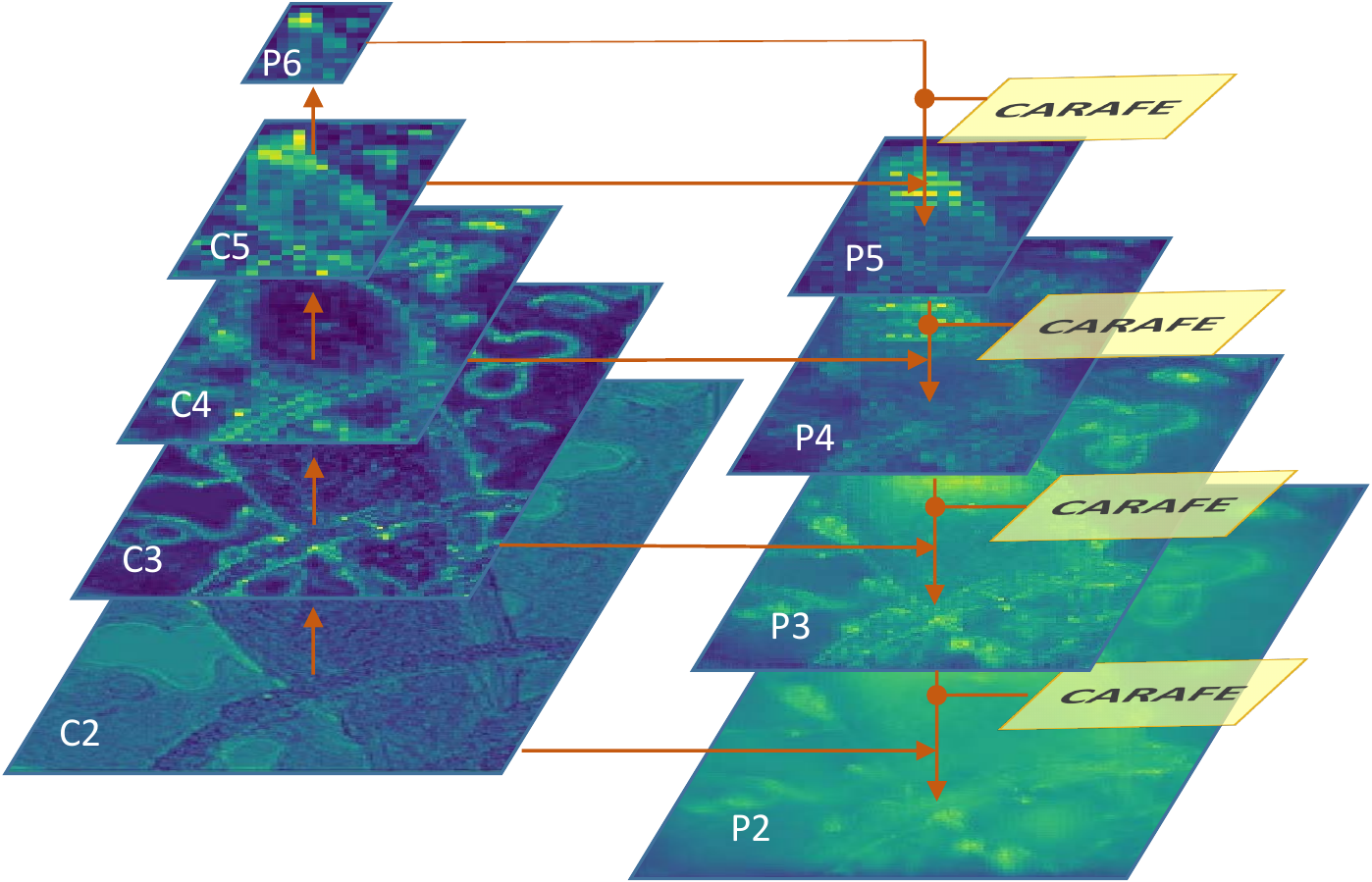}
	\caption{\small{\textbf{FPN architecture with \algname}. \algname~upsamples a feature map by a factor of 2 in the top-down pathway. It is integrated into FPN by seamlessly substituting the nearest neighbor interpolation.}}
	\label{fig:fpn}
	\vspace{-10pt}
\end{figure}

\noindent\textbf{Feature Pyramid Network (FPN).} FPN is an important and effective architecture in the
field of object detection and instance segmentation. It significantly improves
the performance of popular frameworks like Faster R-CNN and Mask R-CNN.
FPN constructs feature pyramids of strong semantics with the top-down pathway
and lateral connections. In the top-down pathway, a low-resolution feature map
is firstly upsampled by 2x with the nearest neighbor interpolation and then fused with a
high-resolution one, as shown in Figure~\ref{fig:fpn}.
We propose to substitute the nearest neighbor interpolation in all the feature levels
with \algname. This modification is smooth and no extra change is required.

\noindent\textbf{Mask Head.} In addition, Mask R-CNN adopts a deconvolution layer at
the end of mask head. It is used to upsample the predicted digits from
$14\times14$ to $28\times28$, to obtain finer mask predictions.
We can also use \algname~to replace the deconvolution layer, resulting in
even less computational cost.

\subsection{Semantic Segmentation}
\label{subsec:application in semantic segmentation}

Semantic segmentation requires the model to output per-pixel level predictions
on the whole image, so that high-resolution feature maps are usually preferred.
In a common pipeline of semantic segmentation model, the pretrained backbone, \eg, ResNet, with downsampling layers are first applied to downsample the feature maps.
And then upsampling is widely adopted to enlarge feature maps from the backbone and fuse the semantic
information of different levels in this task.
UperNet is a strong baseline for semantic segmentation. It uses downsampling in ResNet backbone and upsampling in
the following three components, \ie, PPM, FPN, FUSE. We adopt \algname~instead of their original downsamplers and upsamplers.

\noindent\textbf{ResNet Backbone.} \algname~ is adopted as in Section~\ref{subsec:application in object detection and instance segmentation}, \ie, Res-1 and first building block of Res-3, Res-4, Res-5.

\noindent\textbf{Pyramid Pooling Module (PPM).}
PPM is the key component in PSPNet that hierarchically down-samples an input
feature map into multiple scales $\{1\times1, 2\times2, 3\times3, 6\times6\}$,
and then upsamples them back to the original sizes with bilinear interpolation.
The features are finally fused with the original feature by concatenation.
Since the upsampling ratio is very large, we adopt a two-step strategy with
\algname~as a trade-off between performance and efficiency.
Firstly we upsamples the $\{1\times1, 2\times2, 3\times3, 6\times6\}$
features to half the size of the original feature map with bilinear interpolation,
and then use \algname~to further upsample them by 2x.

\noindent\textbf{Feature Pyramid Network (FPN).}
Similar to detection models, UperNet also adopts FPN to enrich the feature
semantics. It only has four different feature levels \{P2, P3, P4, P5\} with strides \{4, 8, 16, 32\}.
We replace the upsampling operators in the same way as Section~\ref{subsec:application in object detection and instance segmentation}.

\noindent\textbf{Multi-level Feature Fusion (FUSE).}
UperNet introduces a multi-level feature fusion module after the FPN.
It upsamples P3, P4, P5 to the same size as P2 by bilinear interpolation
and then fuses these features from different levels by concatenation.
The process is equivalent to a sequential upsampling-concatenation that first
upsamples P5 to P4 and concatenates them, and then upsamples the concatenated
feature map to P3 and so on. We replace the sequential bilinear upsampling here with \algname.

\subsection{Image Inpainting}
\label{subsec:application in image inpainting}
The U-net architecture is popular among recently proposed image inpainting
methods, such as Global\&Local~\cite{iizuka2017globally} and Partial Conv~\cite{liu2018image}.
We simply replace the strided convolution layers with \algname~followed by convolution layer without stride, and also substitute upsampling layers with \algname~.
As for Partial Conv, we can conveniently keep the mask propagation in
\algname~by updating the mask with our content-aware reassembly kernels.

% !TEX root = ../main.tex

\begin{table*}[t]
	\centering
	\caption{\small{Detection and Instance Segmentation results on MS COCO 2017 \emph{test-dev} with 2x training schedule.}}
	\vspace{-6pt}
	\addtolength{\tabcolsep}{-0pt}
	\small{
		\begin{tabular}{cccccccccc}
			\hline
			\multirow{2}{*}{Method}                 & \multirow{2}{*}{Backbone} & \multirow{2}{*}{Task} & \multirow{2}{*}{AP} & \multirow{2}{*}{$\text{AP}_{50}$} & \multirow{2}{*}{$\text{AP}_{75}$} & \multirow{2}{*}{$\text{AP}_{S}$} & \multirow{2}{*}{$\text{AP}_{M}$} & \multirow{2}{*}{$\text{AP}_{L}$} & \multirow{2}{*}{Inference speed} \\
			                                        &                           &                       &                     &                                   &                                   &                                  &                                  &                                                                     \\ \hline
			Faster R-CNN                            & ResNet-101                & BBox                  & 39.7                & 61.3                              & 43.2                              & 22.0                             & 43.1                             & 50.2                             & 10.3 fps                         \\
			Faster R-CNN w/ \algname                & ResNet-101                & BBox                  & \textbf{42.2}       & \textbf{64.2}                     & \textbf{46.2}                     & \textbf{24.8}                    & \textbf{45.5}                    & \textbf{53.0}                    & 9.9 fps                          \\
			\hline
			\multirow{2}{*}{Mask R-CNN}             & ResNet-101                & BBox                  & 40.8                & 62.3                              & 44.6                              & 22.9                             & 43.9                             & 52.0                             & \multirow{2}{*}{7.6 fps}         \\
			                                        & ResNet-101                & Segm                  & 37.0                & 59.1                              & 39.6                              & 16.9                             & 39.4                             & 53.1                             &                                  \\
			\multirow{2}{*}{Mask R-CNN w/ \algname} & ResNet-101                & BBox                  & \textbf{43.2}       & \textbf{65.1}                     & \textbf{47.4}                     & \textbf{25.8}                    & \textbf{46.4}                    & \textbf{54.4}                    & \multirow{2}{*}{7.3 fps}         \\
			                                        & ResNet-101                & Segm                  & \textbf{39.1}       & \textbf{62.1}                     & \textbf{42.1}                     & \textbf{19.3}                    & \textbf{41.6}                    & \textbf{55.4}                    &                                  \\
			\hline
			Faster R-CNN                            & ResNet-101-DCNv2          & BBox                  & 43.2                & 65.1                              & 47.4                              & 24.6                             & 46.4                             & 55.8                             & 8.5 fps                          \\
			Faster R-CNN w/ \algname                & ResNet-101-DCNv2          & BBox                  & \textbf{44.3}       & \textbf{65.9}                     & \textbf{48.5}                     & \textbf{25.8}                    & \textbf{47.4}                    & \textbf{56.6}                    & 8.3 fps                          \\
			\hline
			\multirow{2}{*}{Mask R-CNN}             & ResNet-101-DCNv2          & BBox                  & 44.2                & 65.8                              & 48.5                              & 25.4                             & 47.3                             & 57.1                             & \multirow{2}{*}{6.7 fps}         \\
			                                        & ResNet-101-DCNv2          & Segm                  & 39.7                & 62.7                              & 42.6                              & 18.5                             & 42.0                             & 57.3                             &                                  \\
			\multirow{2}{*}{Mask R-CNN w/ \algname} & ResNet-101-DCNv2          & BBox                  & \textbf{45.3}       & \textbf{66.7}                     & \textbf{49.7}                     & \textbf{26.8}                    & \textbf{48.7}                    & \textbf{58.0}                    & \multirow{2}{*}{6.5 fps}         \\
			                                        & ResNet-101-DCNv2          & Segm                  & \textbf{40.6}       & \textbf{63.8}                     & \textbf{43.9}                     & \textbf{19.9}                    & \textbf{43.1}                    & \textbf{57.8}                    &                                  \\
			\hline
		\end{tabular}
	}
	\label{tab:det-results}
	\vspace{-10pt}
\end{table*}

\section{Experiments}
\label{sec:experiments}

\subsection{Experimental Settings}

\noindent
\textbf{Datasets \& Evaluation Metrics.}
We evaluate \algname~on several important dense prediction benchmarks.
We use the \emph{train} split for training and evaluate the performance on
the \emph{val} split for all these datasets if it is not further specified. The inference speed is reported on a single TiTan XP GPU.

\noindent\underline{\emph{Image Classification.}}
To evaluate \algname~on the backbone of dense prediction tasks, we pretrain backbones on ImageNet-1k~\cite{deng2009imagenet}~ train split and evaluate the Top-1 and Top-5 accuracy on val split with the single-crop testing.

\noindent\underline{\emph{Object Detection and Instance Segmentation.}}
We perform experiments on the challenging MS COCO 2017~\cite{lin2014microsoft} dataset.
Results are evaluated with the standard COCO metric, \ie AP of IoUs from 0.5 to 0.95.

\noindent\underline{\emph{Semantic Segmentation.}}
We adopt the ADE20k~\cite{zhou2017scene, zhou2018semantic} benchmark to evaluate our method in the semantic segmentation task.
Results are measured with mean IoU (mIoU) and Pixel Accuracy (P.A.),
which respectively indicates the average IoU between predictions and ground
truth masks and per-pixel classification accuracy.

\noindent\underline{\emph{Image Inpainting.}}
Places~\cite{zhou2017places} dataset is adopted for image inpainting.
We use L1 error (lower is better) and PSNR (higher is better) as evaluation metrics.

\noindent
\textbf{Implementation Details.}
If not otherwise specified, \algname~adopts a fixed set of hyper-parameters in
experiments, where $C_m$ is 16 and 64 for the channel compressor in downsampling and upsampling, respectively. And
$k_{encoder}=3$, $k_{reassembly}=5$ for the content encoder.

\noindent\underline{\emph{Image Classification.}}
We evaluate \algname~on ResNet-50 and ResNet-101 backbones. As described in Section~\ref{subsec:application in object detection and instance segmentation},
~\algname~is applied in Bottleneck with strided convolution layers in Res-3, Res-4, Res-5, and replaces the max pooling in Res-1.
Furthermore, a BatchNorm (BN)~and a Relu~layer are adopted after the channel compressor,
which is a 1x1 Conv that compresses input channels. The experimental settings mostly follow~\cite{goyal2017imagenet1hr}.
We use 16 GPUs with batch size of 1024, \ie, 64 images per GPU. We train backbones for 90 epochs in total on ImageNet-1k classification dataset.

\noindent\underline{\emph{Object Detection and Instance Segmentation.}}
We evaluate \algname~on Faster R-CNN~\cite{ren2015faster} and Mask R-CNN~\cite{he2017mask} with the ResNet-50~\cite{he2016deep} w/ FPN~\cite{lin2017feature} backbone for ablation study and report the test-dev results with ResNet-101 backbones.
In both training and inference, we resize an input image such that its shorter edge has 800 pixels or longer edge has 1333 pixels without changing its aspect ratio.
We adopt synchronized SGD with an initial learning rate of 0.02, a momentum of 0.9 and a weight decay of 0.0001. We use a batchsize of 16 over 8 GPUs
(2 images per GPU). Following the 1x and 2x training schedule as Detectron~\cite{Detectron2018} and MMDetection~\cite{mmdetection}.

\begin{figure}[t]
	\centering
	\includegraphics[width=\linewidth]{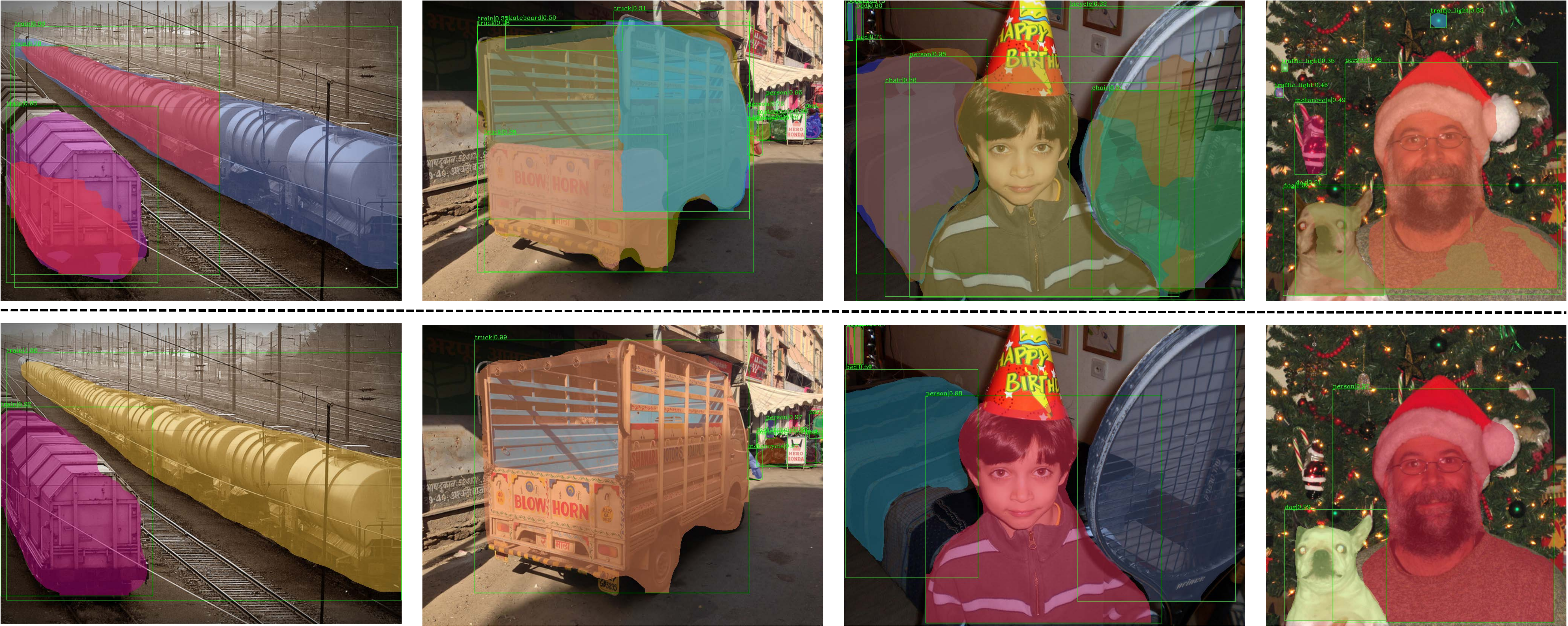}
	\caption{\small{Comparison of instance segmentation results between baseline (top row) and \algname~ (bottom row) on COCO 2017 val.}}
	\label{fig:instance_results}
	\vspace{-10pt}
\end{figure}

\noindent\underline{\emph{Semantic Segmentation.}}
We use the official implementation of UperNet~\cite{xiao2018unified}
with the ResNet backbone as the baseline. During the training, an input image is resized such that the size of its shorter edge is randomly selected from \{300, 375, 450, 525, 600\}. In inference, we apply the single scale
testing for a fair comparison and the shorter edge of an image is set to 450 pixels. The maximum length of the longer edge of an image is set to 1200 in both training and inference.
We adopt synchronized SGD with an initial learning rate of 0.02, a momentum of 0.9 and a weight decay of 0.0001. We use a batchsize of 16 over 8 GPUs
(2 images per GPU), and synchronized batch normalization is adopted as a common practice in semantic segmentation. Following~\cite{Chen_2018_deeplab}, the `poly' learning rate policy is adopted. And we train the model for 20 epochs.

\begin{table}[t]
	\small
	\centering
	\caption{\small{Image classification results on ImageNet-1k val.}}
	\vspace{-6pt}
	\addtolength{\tabcolsep}{-2pt}
	\small{
		\begin{tabular}{ccc}
			\hline
			\multirow{2}{*}{Model} & \multirow{2}{*}{Top-1} & \multirow{2}{*}{Top-5} \\ \
			                       &                        &                        \\ \hline
			ResNet-50              & 76.43                  & 93.21                  \\
			ResNet-50 w/ CARAFE++  & \textbf{77.46}         & 93.63                  \\ \hline
			ResNet-101             & 78.10                  & 93.92                  \\
			ResNet-101 w/ CARAFE++ & \textbf{78.70}         & 94.31                  \\ \hline
		\end{tabular}
	}
	\label{tab:downsample_classification}
	\vspace{-5pt}
\end{table}

\begin{table}[t]
	\centering
	\caption{\small{Detection results with Faster R-CNN. Various downsampling methods are used in ResNet-50 backbone. `Conv', `MPool', `APool', 'DPP', 'LIP' indicates convolution layer with stride, Max Pooling, Average Pooling, Detail Preserved Pooling~\cite{Saeedan_2018_CVPR}, and Local Importance-based Pooling~\cite{Gao_2019_ICCV}, respectively. The 1x training schedule is applied in experiments.}}
	\vspace{-6pt}
	\addtolength{\tabcolsep}{-3pt}
	\small{
		\begin{tabular}{ccccccccc}
			\hline
			\multirow{2}{*}{Method}      & \multirow{2}{*}{$\text{AP}_{box}$} & \multirow{2}{*}{$\text{AP}_{50}$} & \multirow{2}{*}{$\text{AP}_{75}$} & \multirow{2}{*}{$\text{AP}_{S}$} & \multirow{2}{*}{$\text{AP}_{M}$} & \multirow{2}{*}{$\text{AP}_{L}$} & \multirow{2}{*}{FPS} \\
			                             &                                    &                                   &                                   &                                  &                                  &                                  &                      \\ \hline
			Conv                         & 36.5                               & 58.4                              & 39.3                              & 21.3                             & 40.3                             & 47.2                             & 12.5 fps             \\
			Max Pool                     & 37.8                               & 59.6                              & 40.9                              & 22.3                             & 42.1                             & 48.4                             & 12.5 fps             \\
			Avg Pool                     & 37.9                               & 59.8                              & 40.9                              & 22.1                             & 42.1                             & 47.9                             & 12.3 fps             \\
			DPP~\cite{Saeedan_2018_CVPR} & 37.7                               & 59.7                              & 40.6                              & 22.3                             & 41.9                             & 48.1                             & 10.9 fps             \\
			LIP~\cite{Gao_2019_ICCV}     & 38.0                               & 60.3                              & 41.2                              & 23.1                             & 41.9                             & 48.2                             & 10.8 fps             \\
			\algname                     & \textbf{38.8}                      & 60.8                              & 42.2                              & 23.3                             & 43.1                             & 49.2                             & 12.1 fps             \\
			\hline
		\end{tabular}
	}
	\label{tab:downsampl_compare}
	\vspace{-5pt}

\end{table}

\noindent\underline{\emph{Image Inpainting.}}
We adopt Global\&Local~\cite{iizuka2017globally} as the baseline method to evaluate \algname.
We employ the generator and discriminator networks from Global\&Local~\cite{iizuka2017globally}.
Our generator takes a $256 \times 256$ image $\mathbf{x}$ with masked region $M$ as input and produces a $256 \times 256$ prediction of the missing region $\mathbf{\hat{y}}$ as output.
Then we combine the predicted image with the input by $\mathbf{y}=(1-M)\odot \mathbf{x} + M \odot \mathbf{\hat{y}}$.
Finally, the combined output $\mathbf{y}$ is fed into the discriminator.
We apply a simple modification to the baseline model to achieve better generation quality.
Compared to the original model that employs two discriminators, we employ only one PatchGAN-style discriminator\cite{li2016precomputed} on the inpainted region.
For a fair comparison, we use the free-form masks introduced by \cite{yu2018free} as the binary mask $M$.
We further compare \algname with Partial Conv~\cite{liu2018image}, we substitute the convolution layers with the official Partial Conv module in our generator.
During training, Adam solver with learning rate 0.0001 is adopted where $\beta_1=0.5$ and $\beta_2=0.9$. Training batch size is 32.
The input and output are linearly scaled within range $[-1, 1]$.

\begin{table}[t]
	\centering
	\caption{\small{Detection results with Faster R-CNN. Various upsampling methods are used in FPN. N.C., B.C., P.S. and S.A.
			indicate Nearest + Conv, Bilinear + Conv, Pixel Shuffle and Spatial Attention, respectively.}}
	\vspace{-6pt}
	\addtolength{\tabcolsep}{-3pt}
	\small{
		\begin{tabular}{ccccccccccc}
			\hline
			\multirow{2}{*}{Method}     & \multirow{2}{*}{$\text{AP}_{box}$} & \multirow{2}{*}{$\text{AP}_{50}$} & \multirow{2}{*}{$\text{AP}_{75}$} & \multirow{2}{*}{$\text{AP}_{S}$} & \multirow{2}{*}{$\text{AP}_{M}$} & \multirow{2}{*}{$\text{AP}_{L}$} & \multirow{2}{*}{FPS} \\
			                            &                                    &                                   &                                   &                                  &                                  &                                  &                      \\ \hline
			Nearest                     & 36.5                               & 58.4                              & 39.3                              & 21.3                             & 40.3                             & 47.2                             & 12.5 fps             \\
			Bilinear                    & 36.7                               & 58.7                              & 39.7                              & 21.0                             & 40.5                             & 47.5                             & 12.5 fps             \\
			N.C.                        & 36.6                               & 58.6                              & 39.5                              & 21.4                             & 40.3                             & 46.4                             & 11.8 fps             \\
			B.C.                        & 36.6                               & 58.7                              & 39.4                              & 21.6                             & 40.6                             & 46.8                             & 11.7 fps             \\
			Deconv~\cite{Noh_2015}      & 36.4                               & 58.2                              & 39.2                              & 21.3                             & 39.9                             & 46.5                             & 11.5 fps             \\
			P.S.\cite{Shi_2016}         & 36.5                               & 58.8                              & 39.1                              & 20.9                             & 40.4                             & 46.7                             & 11.7 fps             \\
			GUM\cite{mazzini2018guided} & 36.9                               & 58.9                              & 39.7                              & 21.5                             & 40.6                             & 48.1                             & 11.8 fps             \\
			S.A.\cite{chen2017sca}      & 36.9                               & 58.8                              & 39.8                              & 21.7                             & 40.8                             & 47.0                             & 12.3 fps             \\
			\algname~                   & \textbf{37.8}                      & 60.1                              & 40.8                              & 23.2                             & 41.2                             & 48.2                             & 12.2 fps             \\ \hline
		\end{tabular}
	}
	\label{tab:upsample compare}
	\vspace{-5pt}

\end{table}

\begin{table}[t]
	\centering
	\caption{\small{Object detection results of Faster R-CNN with \algname~in FPN and backbone. }}
	\vspace{-6pt}
	\addtolength{\tabcolsep}{-3pt}
	\small{
		\begin{tabular}{ccccccccc}
			\hline
			\multirow{2}{*}{FPN} & \multirow{2}{*}{Backbone} & \multirow{2}{*}{AP} & \multirow{2}{*}{$\text{AP}_{50}$} & \multirow{2}{*}{$\text{AP}_{75}$} & \multirow{2}{*}{$\text{AP}_{S}$} & \multirow{2}{*}{$\text{AP}_{M}$} & \multirow{2}{*}{$\text{AP}_{L}$} & \multirow{2}{*}{FPS} \\
			                     &                           &                     &                                   &                                   &                                  &                                  &                                                         \\ \hline
			\checkmark           &                           & 37.8                & 60.1                              & 40.8                              & 23.2                             & 41.2                             & 48.2                             & 12.2 fps             \\
			                     & \checkmark                & 38.8                & 60.8                              & 42.2                              & 23.3                             & 43.1                             & 49.2                             & 12.1 fps             \\
			\checkmark           & \checkmark                & \bf{39.6}           & 62.0                              & 43.2                              & 24.7                             & 43.7                             & 50.9                             & 11.7 fps             \\ \hline
		\end{tabular}
	}
	\label{tab:faster rcnn}
	\vspace{-5pt}

\end{table}

\begin{table}[t]
	\centering
	\caption{\small{Instance Segmentation results with Mask R-CNN. Various upsampling methods are used in mask head.}}
	\vspace{-6pt}
	\addtolength{\tabcolsep}{-2pt}
	\small{
		\begin{tabular}{ccccccc}
			\hline
			\multirow{2}{*}{Method} & \multirow{2}{*}{AP} & \multirow{2}{*}{$\text{AP}_{50}$} & \multirow{2}{*}{$\text{AP}_{75}$} & \multirow{2}{*}{$\text{AP}_{S}$} & \multirow{2}{*}{$\text{AP}_{M}$} & \multirow{2}{*}{$\text{AP}_{L}$} \\
			                        &                     &                                   &                                   &                                  &                                  &                                  \\ \hline
			Nearest                 & 32.7                & 55.0                              & 34.8                              & 17.7                             & 35.9                             & 44.4                             \\
			Bilinear                & 34.2                & 55.9                              & 36.4                              & 18.5                             & 37.5                             & 46.2                             \\
			Deconv                  & 34.2                & 55.5                              & 36.3                              & 17.6                             & 37.8                             & 46.7                             \\
			Pixel Shuffle           & 34.4                & 56.0                              & 36.6                              & 18.5                             & 37.6                             & 47.5                             \\
			GUM                     & 34.3                & 55.7                              & 36.5                              & 17.6                             & 37.6                             & 46.9                             \\
			S.A.                    & 34.1                & 55.6                              & 36.5                              & 17.6                             & 37.4                             & 46.6                             \\
			\algname~               & \textbf{34.7}       & 56.2                              & 37.1                              & 18.2                             & 37.9                             & 47.5                             \\ \hline
		\end{tabular}
	}
	\label{tab:mask head compare}
	\vspace{-5pt}

\end{table}

\begin{table}[ht]
	\centering
	\caption{\small{Detection and Instance Segmentation results of Mask R-CNN with \algname~in Backbone, FPN and mask head, respectively.
			M.H. indicates using \algname~in mask head. }}
	\vspace{-6pt}
	\addtolength{\tabcolsep}{-4pt}
	\small{
		\begin{tabular}{cccccccccc}
			\hline
			\multirow{2}{*}{FPN}        & \multirow{2}{*}{M.H.}       & \multirow{2}{*}{B.K.}       & \multirow{2}{*}{Task} & \multirow{2}{*}{AP} & \multirow{2}{*}{$\text{AP}_{50}$} & \multirow{2}{*}{$\text{AP}_{75}$} & \multirow{2}{*}{$\text{AP}_{S}$} & \multirow{2}{*}{$\text{AP}_{M}$} & \multirow{2}{*}{$\text{AP}_{L}$} \\
			                            &                             &                             &                       &                     &                                   &                                   &                                  &                                  &                                  \\ \hline
			\multirow{2}{*}{}           & \multirow{2}{*}{}           & \multirow{2}{*}{}           & Bbox                  & 37.4                & 59.1                              & 40.3                              & 21.2                             & 41.2                             & 48.5                             \\
			                            &                             & \multirow{2}{*}{}           & Segm                  & 34.2                & 55.5                              & 36.3                              & 17.6                             & 37.8                             & 46.7                             \\ \hline
			\multirow{2}{*}{\checkmark} & \multirow{2}{*}{}           & \multirow{2}{*}{}           & Bbox                  & 38.6                & 60.7                              & 42.2                              & 23.2                             & 42.1                             & 49.5                             \\
			                            &                             & \multirow{2}{*}{}           & Segm                  & 35.2                & 57.2                              & 37.5                              & 19.3                             & 38.3                             & 47.6                             \\ \hline
			\multirow{2}{*}{}           & \multirow{2}{*}{\checkmark} & \multirow{2}{*}{}           & Bbox                  & 37.3                & 59.0                              & 40.2                              & 21.8                             & 40.8                             & 48.6                             \\
			                            &                             & \multirow{2}{*}{}           & Segm                  & 34.7                & 56.2                              & 37.1                              & 18.2                             & 37.9                             & 47.5                             \\ \hline
			\multirow{2}{*}{\checkmark} & \multirow{2}{*}{\checkmark} & \multirow{2}{*}{}           & Bbox                  & 38.6                & 60.6                              & 42.1                              & 23.2                             & 42.1                             & 50.0                             \\
			                            &                             & \multirow{2}{*}{}           & Segm                  & 35.6                & 57.5                              & 37.8                              & 16.9                             & 38.1                             & 52.4                             \\ \hline
			\multirow{2}{*}{}           & \multirow{2}{*}{}           & \multirow{2}{*}{\checkmark} & Bbox                  & 39.5                & 61.3                              & 43.2                              & 23.9                             & 43.4                             & 50.7                             \\
			                            &                             & \multirow{2}{*}{}           & Segm                  & 35.9                & 57.9                              & 38.0                              & 17.5                             & 38.9                             & 51.8                             \\ \hline
			\multirow{2}{*}{\checkmark} & \multirow{2}{*}{\checkmark} & \multirow{2}{*}{\checkmark} & Bbox                  & \bf{40.3}           & 62.5                              &
			43.9                        & 24.8                        & 44.4                        & 51.5                                                                                                                                                                                                                         \\
			                            &                             & \multirow{2}{*}{}           & Segm                  & \bf{36.8}           & 59.2                              & 39.2                              & 18.0                             & 39.9                             & 53.6                             \\ \hline
		\end{tabular}
		\vspace{-5pt}
	}
	\label{tab:mask rcnn}

\end{table}

\subsection{Benchmarking Results}
\label{subsec:results}

\noindent
\textbf{Object Detection \& Instance Segmentation.}
We first evaluate ~\algname~in Faster R-CNN and Mask R-CNN to show its effectiveness in object detection and instance segmentation. We first replace the downsampling layers in ResNet-101 backbone with \algname~and pretrain the modified backbone in ImageNet-1k classification dataset as in Section~\ref{subsec:application in object detection and instance segmentation}. We then substitute the nearest neighbor interpolation in FPN with
\algname~for both Faster R-CNN and Mask R-CNN, and the deconvolution layer in the
mask head for Mask R-CNN.
As shown in Table~\ref{tab:det-results}, with ResNet-101 backbone and 2x training schedule, \algname~improves Faster R-CNN by 2.5\% (\ie, from 39.7\% to 42.2\%)
on $AP_{box}$, and Mask R-CNN by 2.1\% (\ie, from 37.0\% to 39.1\%) on $AP_{mask}$ with minor extra inference time (\ie, 10.3 fps v.s. 9.9 fps and 7.6 fps v.s. 7.3 fps).
The results show the effectiveness and efficiency of \algname. In Figure~\ref{fig:instance_results}, we show some examples of instance
segmentation results comparing the baseline and \algname.

We further apply \algname~in ResNet-101 backbone with powerful Deformable Convolution layers v2 (DCNv2)~\cite{Zhu_2019_CVPR}. In the ResNet-101 with DCNv2 baseline, the 3x3 Convolution layers in Res-2, Res-3, Res-4 are replaced with 3x3 DCNv2. To investigate the effectiveness of \algname, DCNv2 layers with stride of 2 are replaced with \algname~and a following 3x3 convolution layers without stride. As shown in Table~\ref{tab:det-results}, \algname~still achieves substantial gains on such a strong baseline. Specifically, \algname improve $\sim$1\% $AP$ for both Faster R-CNN and Mask R-CNN with similar inference speed.

\noindent\underline{\emph{\algname~for Downsampling.}} Here we investigate the effectiveness of \algname~for downsampling.
As illustrated in Table~\ref{tab:downsample_classification}, after integrated into ResNet backbone, \algname~improves the Top-1 accuracy of image classification on ResNet-50 and ResNet-100 by 1\% and 0.6\% respectively. Therefore, apart from dense predication tasks, \algname~also brings benefits to the classification task.

We further explore different downsampling methods in object detection. To be specific, we adopt different downsampling methods in the ResNet-50 backbone, and then pretrain the backbone on ImageNet-1k classification dataset. The pretrained backbones are used to train Faster R-CNN with FPN on COCO dataset. As summarized in Table~\ref{tab:downsampl_compare}, we extensively compare convolution layer with stride, Max Pooling, Average Pooling, Detail Preserved Pooling~\cite{Saeedan_2018_CVPR}, Local Importance-based Pooling~\cite{Gao_2019_ICCV} and \algname.
The strided convolution layer is the baseline. For Max Pooling, Average Pooling, Detail Preserved Pooling, we apply these pooling methods in the same manner with \algname~as described in Section~\ref{subsec:application in object detection and instance segmentation}. LIP~\cite{Gao_2019_ICCV} is adopted following its original paper.
We observe that by simply replacing the baseline (\ie, strided convolution layers) with Max/Average/Detail-Preserved pooling, the performance can be improved by $\sim$1.3\%. The recent proposed LIP achieves a slightly better results than these pooling operators. \algname~significantly improves the baseline by 2.3\% (\ie, from 36.5\% to 38.8\% AP) with minor extra cost (\ie, from 12.5 fps to 12.1 fps).

\noindent\underline{\emph{\algname~for Upsampling.}} To investigate the effectiveness of different upsampling operators, we perform
extensive experiments on Faster R-CNN by using different operators to perform upsampling in FPN.
Results are illustrated in Table~\ref{tab:upsample compare}.
For `N.C.' and `B.C.', which respectively indicate `Nearest + Conv' and `Bilinear + Conv', we add an extra $3\times3$
convolution layer after the corresponding upsampling.
`Deconv', `Pixel Shuffle' (indicated as `P.S.'), `GUM' are three representative learning-based upsampling methods.
We also compare `Spatial Attention' here, indicated as `S.A.'.
\algname~achieves the best $AP_{box}$ among all these upsampling operators with minor extra computation cost, which illustrates it is both effective and efficient.
The results of `Nearest + Conv' and `Bilinear + Conv' show that simply adding more convolution layers
does not lead to a significant gain. However, it slows down the inference speed from 12.5 fps to 11.8 fps.
`Deconv', `Pixel Shuffle', `GUM' and `Spatial Attention' obtain inferior performance to \algname,
indicating that the design of effective upsampling operators is critical.

In Table~\ref{tab:faster rcnn}, we report the object detection
performance of adopting \algname~in Backbone and FPN, respectively.
Adopting \algname~ both in Backbone and FPN further improves the object detection performance to 39.6\%.

\noindent\underline{\emph{\algname~for Mask Predication.}} Besides FPN, which is a pyramid feature fusion structure, we also explore different
upsampling operators in the mask head.
In typical Mask R-CNN, a deconvolution layer is adopted to upsample the RoI
features by 2x.
For a fair comparison, we do not make any changes to FPN, and only replace the
deconvolution layer with various operators.
Since we only modify the mask prediction branch, performance is reported in
terms of $AP_{mask}$, as shown in Table~\ref{tab:mask head compare}.
\algname~achieves the best performance in instance segmentation among these methods.

\begin{figure}[t]
	\centering
	\includegraphics[width=\linewidth]{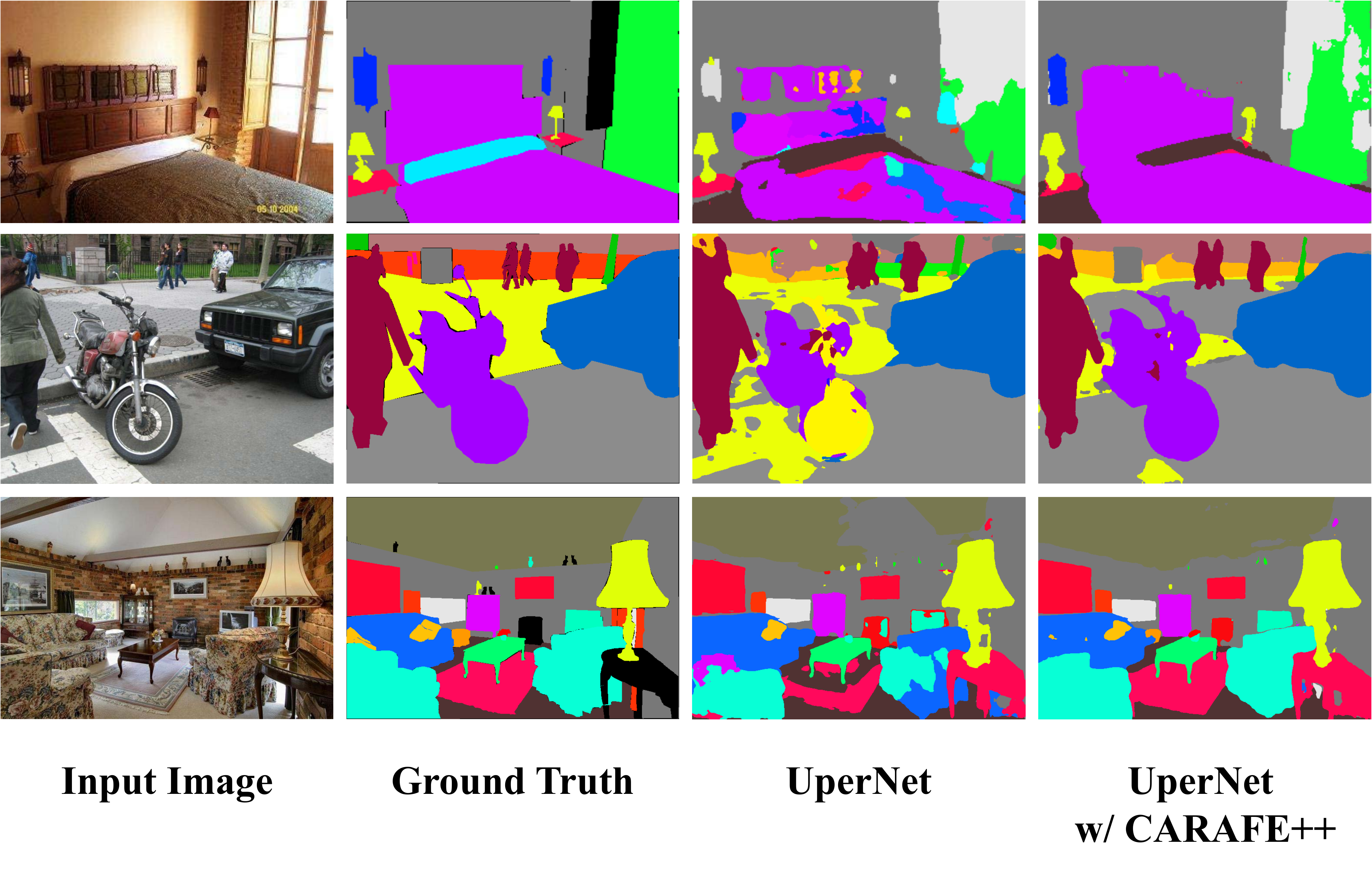}
	\vspace{-21pt}
	\caption{\small{Comparison of semantic segmentation results between UperNet baseline and UperNet w/ \algname~ on ADE20k.}}
	\label{fig:seg_results}
	\vspace{-5pt}
\end{figure}

\begin{table}[t]
	\centering
	\caption{\small{Semantic Segmentation results on ADE20k val. Single scale testing is
			used in our experiments. P.A. indicates Pixel Accuracy.}}
	\vspace{-6pt}
	\addtolength{\tabcolsep}{-2pt}
	\small{
		\begin{tabular}{cccc}
			\hline
			\multirow{2}{*}{Method} & \multirow{2}{*}{Backbone} & \multirow{2}{*}{mIoU} & \multirow{2}{*}{P.A.} \\
			                        &                           &                       &                       \\ \hline
			PSPNet                  & ResNet-50                 & 41.68                 & 80.04                 \\
			PSANet                  & ResNet-50                 & 41.92                 & 80.17                 \\
			UperNet\tablefootnote{We report the performance in model zoo of the official implementation:        \\ \url{https://github.com/CSAILVision/semantic-segmentation-pytorch}}{} & ResNet-50                 & 40.44                 & 79.80                 \\
			UperNet w/ \algname     & ResNet-50                 & \textbf{43.05}        & \textbf{80.81}        \\ \hline
			PSPNet                  & ResNet-101                & 41.96                 & 80.64                 \\
			PSANet                  & ResNet-101                & 42.75                 & 80.71                 \\
			UperNet                 & ResNet-101                & 42.00                 & 80.79                 \\
			UperNet w/ \algname     & ResNet-101                & \textbf{43.94}        & \textbf{81.26}        \\ \hline
		\end{tabular}
	}
	\label{tab:semantic segmentation}
	\vspace{-5pt}

\end{table}

In Table~\ref{tab:mask rcnn}, we report the object detection and instance segmentation
performance of adopting \algname~in Backbone, FPN and mask head on Mask R-CNN, respectively.
Consistent improvements are achieved in these experiments.

\noindent
\textbf{Semantic Segmentation.}
We replace the downsamplers and upsamplers in UperNet with \algname~and evaluate the results on ADE20k benchmark.
As shown in Table~\ref{tab:semantic segmentation}, adopting single scale testing, \algname~improves the mIoU
by a large margin from 40.44\% to 43.05\% on ResNet-50 Backbone and from 42.0\% to 43.94\% with ResNet-101 backbone.
Note that UperNet with \algname~also achieves better performance than other strong baselines such as PSPNet\cite{zhao2017pyramid} and PSANet\cite{zhao2018psanet}.

\begin{figure}[t]
	\centering
	\includegraphics[width=\linewidth]{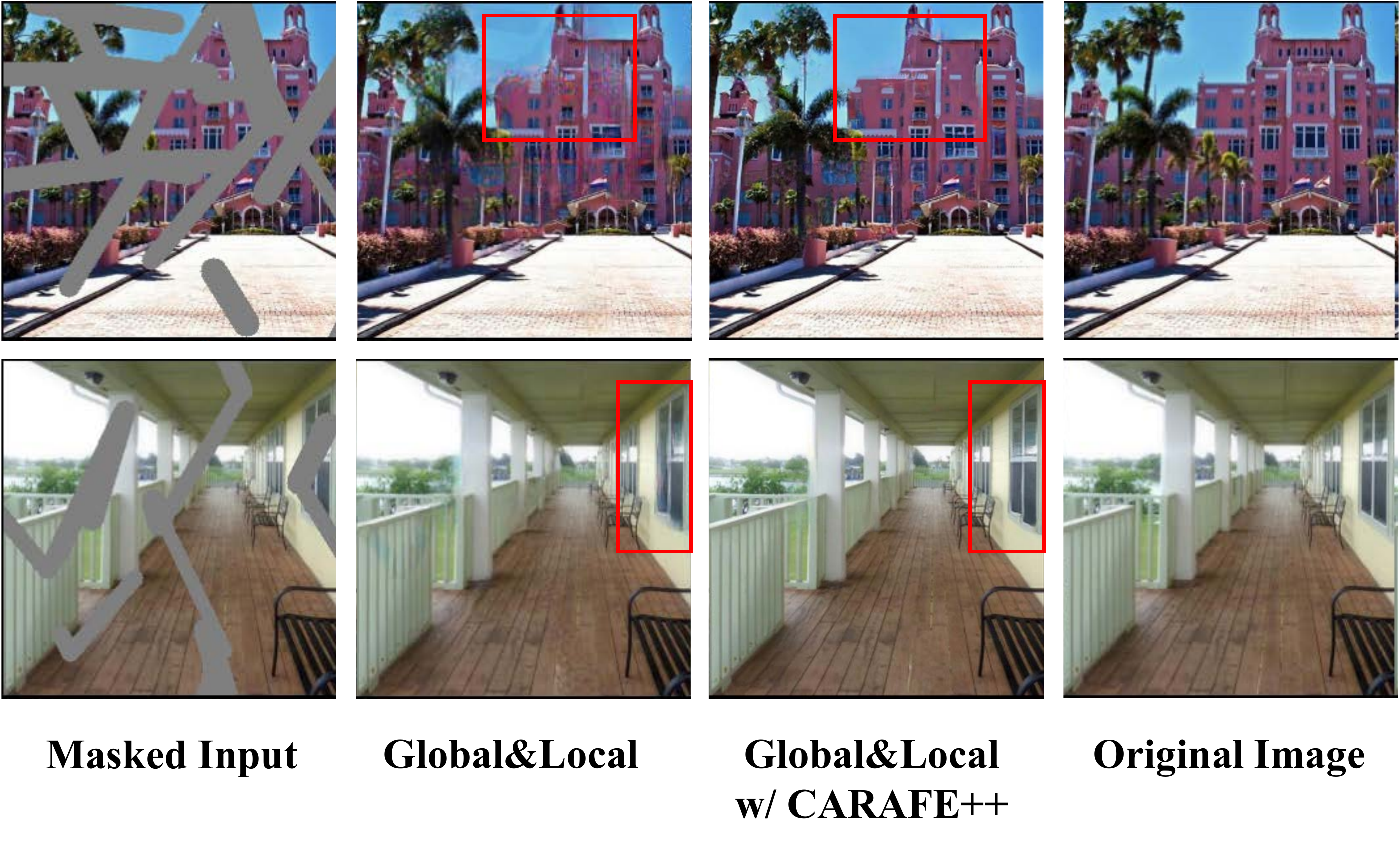}
	\vspace{-21pt}
	\caption{\small{Comparison of image inpainting results between Gobal\&Local baseline and Gobal\&Local w/ \algname~ on ADE20k.}}
	\label{fig:inpaint_results}
	\vspace{-5pt}
\end{figure}

We perform a step-by-step study to inspect the effectiveness of modifying different
components in UperNet, as described in Section~\ref{subsec:application in semantic segmentation}.
Results in Table~\ref{tab:semantic segmentation by step} show that \algname~is
helpful for all the four components and the combination of them results in
further gains. Visualization comparisons of UperNet with and without \algname~are shown in Figure~\ref{fig:seg_results}.

\begin{table}[t]
	\centering
	\caption{\small{Effects of adopting \algname~in each component of UperNet.}}
	\vspace{-5pt}
	\addtolength{\tabcolsep}{-2pt}
	\small{
		\begin{tabular}{cccccc}
			\hline
			\multirow{2}{*}{PPM} & \multirow{2}{*}{FPN} & \multirow{2}{*}{FUSE} & \multirow{2}{*}{Backbone} & \multirow{2}{*}{mIoU} & \multirow{2}{*}{P.A.} \\
			                     &                      &                       &                           &                       &                       \\ \hline
			\checkmark           &                      &                       &                           & 40.85                 & 79.97                 \\
			                     & \checkmark           &                       &                           & 40.79                 & 80.01                 \\
			                     &                      & \checkmark            &                           & 41.06                 & 80.23                 \\
			\checkmark           & \checkmark           &                       &                           & 41.55                 & 80.30                 \\
			\checkmark           &                      & \checkmark            &                           & 42.01                 & 80.11                 \\
			                     & \checkmark           & \checkmark            &                           & 41.93                 & 80.34                 \\
			\checkmark           & \checkmark           & \checkmark            &                           & 42.23                 & 80.34                 \\
			                     &                      &                       & \checkmark                & 41.93                 & 80.35                 \\
			\checkmark           & \checkmark           & \checkmark            & \checkmark                & \bf{43.05}            & \bf{80.81}            \\ \hline
		\end{tabular}
	}
	\label{tab:semantic segmentation by step}
	\vspace{-5pt}
\end{table}

\begin{table}[t]
	\small
	\centering
	\caption{\small{Image inpainting results on Places val.}}
	\vspace{-6pt}
	\addtolength{\tabcolsep}{-2pt}
	\small{
		\begin{tabular}{lll}
			\hline
			\multirow{2}{*}{Method}   & \multirow{2}{*}{L1(\%)} & \multirow{2}{*}{PSNR(dB)} \\
			                          &                         &                           \\ \hline
			Global\&Local             & 6.69                    & 19.58                     \\
			Global\&Local w/ \algname & 5.82                    & 20.93                     \\
			\hline
			Partial Conv              & 5.96                    & 20.78                     \\
			Partial Conv w/ \algname  & \bf{5.60}               & \bf{21.05}                \\ \hline
		\end{tabular}
	}
	\label{tab:inpainting}
	\vspace{-5pt}
\end{table}

\noindent
\textbf{Image Inpainting.}
We show that \algname~is also effective in low-level tasks such as image inpainting.
By replacing the upsampling operators with \algname~in two strong baselines
Global\&Local~\cite{iizuka2017globally} and Partial Conv~\cite{liu2018image}, we observe substantial
improvements for both methods.
As shown in Table~\ref{tab:inpainting}, our method improves two baselines by
1.35 dB and 0.27 dB on the PSNR metric. Visualization results of \algname~are shown in Figure~\ref{fig:inpaint_results}.

\begin{figure*}
	\centering
	\includegraphics[width=\linewidth]{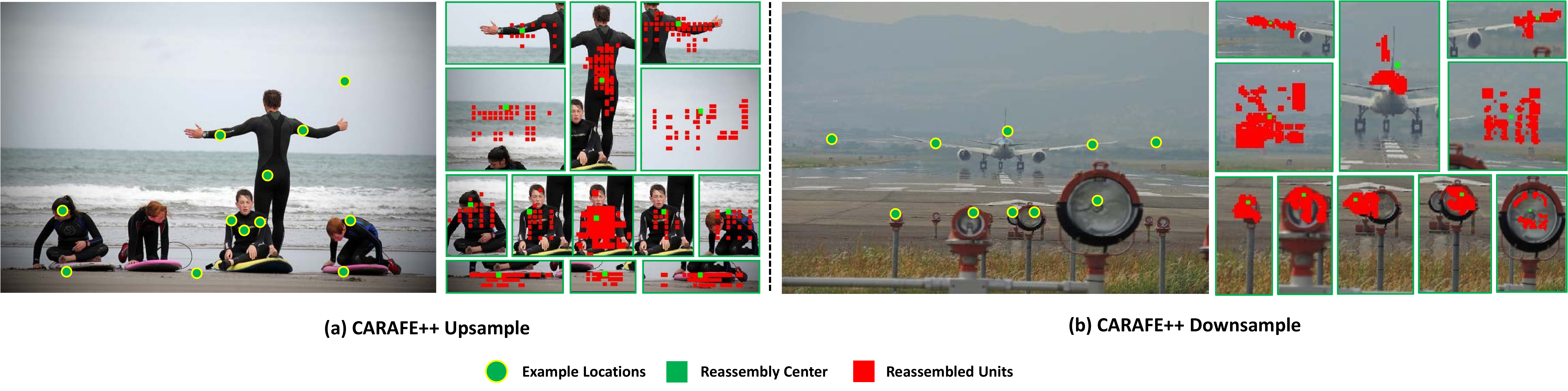}
	\vspace{-21pt}
	\caption{\small{\algname~performs content-aware reassembly when rescaling a feature map.
			Red units are reassembled into the green center unit by \algname. }}
	\vspace{-10pt}
	\label{fig:arrow}
\end{figure*}

\subsection{Ablation Study \& Further Analysis}
\label{subsec:ablation}

\begin{table}[t]
	\centering
	\caption{\small{Ablation study of various compressed channels $C_m$ for upsampling.
			The experiments are performed with Faster R-CNN w/ \algname in FPN structure on COCO dataset.
			N/A means \emph{channel compressor} is removed. }}
	\vspace{-6pt}
	\addtolength{\tabcolsep}{-2pt}
	\small{
		\begin{tabular}{lllllll}
			\hline
			\multirow{2}{*}{$C_m$} & \multirow{2}{*}{AP} & \multirow{2}{*}{$\text{AP}_{50}$} & \multirow{2}{*}{$\text{AP}_{75}$} & \multirow{2}{*}{$\text{AP}_{S}$} & \multirow{2}{*}{$\text{AP}_{M}$} & \multirow{2}{*}{$\text{AP}_{L}$} \\
			                       &                     &                                   &                                   &                                  &                                  &                                  \\ \hline
			16                     & 37.6                & 60.1                              & 40.6                              & 22.7                             & 41.6                             & 48.4                             \\
			32                     & 37.7                & 60.3                              & 40.7                              & 22.8                             & 41.2                             & 49.0                             \\
			64                     & \bf{37.8}           & 60.1                              & 40.8                              & 23.2                             & 41.2                             & 48.2                             \\
			128                    & \bf{37.8}           & 60.1                              & 40.8                              & 22.4                             & 41.7                             & 48.7                             \\
			256                    & \bf{37.8}           & 60.4                              & 40.8                              & 22.7                             & 41.3                             & 48.8                             \\
			N/A                    & \bf{37.8}           & 60.3                              & 40.8                              & 22.9                             & 41.5                             & 48.7                             \\ \hline
		\end{tabular}
		\vspace{-5pt}
	}
	\label{tab:mid channels}

\end{table}

\begin{table}[t]
	\centering
	\caption{\small{Ablation study of various compressed channels $C_m$ for downsampling. The experiments are performed with ResNet-50 on ImageNet-1k classification dataset.}}
	\vspace{-6pt}
	\addtolength{\tabcolsep}{-3pt}
	\small{
		\begin{tabular}{ccc}
			\hline
			\multirow{2}{*}{$C_m$} & \multirow{2}{*}{Top-1} & \multirow{2}{*}{Top-5} \\
			                       &                        &                        \\ \hline
			4                      & 77.31                  & 93.59                  \\
			8                      & 77.43                  & 93.63                  \\
			16                     & \bf{77.46}             & 93.63                  \\
			32                     & 77.40                  & 93.60                  \\
			64                     & \bf{77.46}             & 93.67                  \\ \hline
		\end{tabular}
		\vspace{-5pt}
	}
	\label{tab:mid-cls}

\end{table}

\begin{table}[t]
	\centering
	\caption{\small{Detection results with various encoder kernel size $k_{encoder}$ and reassembly
			kernel size $k_{reassembly}$.} }
	\vspace{-6pt}
	\addtolength{\tabcolsep}{-2pt}
	\small{
		\begin{tabular}{llllllll}
			\hline
			\multirow{2}{*}{$k_{encoder}$} & \multirow{2}{*}{$k_{reassembly}$} & \multirow{2}{*}{AP} & \multirow{2}{*}{$\text{AP}_{50}$} & \multirow{2}{*}{$\text{AP}_{75}$} & \multirow{2}{*}{$\text{AP}_{S}$} & \multirow{2}{*}{$\text{AP}_{M}$} & \multirow{2}{*}{$\text{AP}_{L}$} \\
			                               &                                   &                     &                                   &                                   &                                  &                                  &                                  \\ \hline
			1                              & 3                                 & 37.3                & 59.6                              & 40.5                              & 22.0                             & 40.7                             & 48.1                             \\
			1                              & 5                                 & 37.3                & 59.9                              & 40.0                              & 22.3                             & 41.1                             & 47.3                             \\
			3                              & 3                                 & 37.3                & 59.7                              & 40.4                              & 22.1                             & 40.8                             & 48.3                             \\
			3                              & 5                                 & 37.8                & 60.1                              & 40.8                              & 23.2                             & 41.2                             & 48.2                             \\
			3                              & 7                                 & 37.7                & 60.0                              & 40.9                              & 23.0                             & 41.5                             & 48.4                             \\
			5                              & 5                                 & 37.8                & 60.2                              & 40.7                              & 22.5                             & 41.4                             & 48.6                             \\
			5                              & 7                                 & \bf{38.1}           & 60.4                              & 41.3                              & 23.0                             & 41.6                             & 48.8                             \\
			7                              & 7                                 & 38.0                & 60.2                              & 41.1                              & 23.0                             & 41.8                             & 48.8                             \\ \hline
		\end{tabular}
		\vspace{-5pt}
	}
	\label{tab:k study}

\end{table}

\begin{table}[t]
	\centering
	\caption{\small{Ablation study of different normalization methods in \emph{kernel normalizer}.} }
	\vspace{-6pt}
	\addtolength{\tabcolsep}{-2pt}
	\small{
		\begin{tabular}{lllllll}
			\hline
			\multirow{2}{*}{Method} & \multirow{2}{*}{AP} & \multirow{2}{*}{$\text{AP}_{50}$} & \multirow{2}{*}{$\text{AP}_{75}$} & \multirow{2}{*}{$\text{AP}_{S}$} & \multirow{2}{*}{$\text{AP}_{M}$} & \multirow{2}{*}{$\text{AP}_{L}$} \\
			                        &                     &                                   &                                   &                                  &                                  &                                  \\ \hline
			Sigmoid                 & 37.4                & 59.8                              & 40.2                              & 23.1                             & 40.9                             & 47.4                             \\
			Sigmoid Normalize       & \bf{37.8}           & 60.1                              & 40.7                              & 22.6                             & 41.6                             & 48.0                             \\
			Softmax                 & \bf{37.8}           & 60.1                              & 40.8                              & 23.2                             & 41.2                             & 48.2                             \\ \hline
		\end{tabular}
		\vspace{-10pt}
	}
	\label{tab:softmax}

\end{table}

\noindent
\textbf{Model Design \& Hyper-parameters.}
We investigate the influence of hyper-parameters in the model design, \ie,
the compressed channels $C_m$, encoder kernel size $k_{encoder}$ and reassembly
kernel size $k_{reassembly}$. We also test different normalization methods in the
kernel normalizer.
Faster R-CNN with a ResNet-50 backbone and 1x schedule is adopted for ablation study if not further specified.

We explore different values of $C_m$ in the channel compressor. The experiments are conducted on object detection for upsampling and image classification for downsampling.
For the upsampling version of \algname~, the influences of $C_m$ is evaluated on Faster R-CNN with ResNet-50 backbone.
Experimental results in Table~\ref{tab:mid channels} show that compress $C_m$
down to 64 leads to no performance decline, while being more efficient.
A further smaller $C_m$ will result in a slight drop of the performance.
In addition, we also try removing the channel compressor module, which means
the content encoder directly uses input features to predict reassembly kernels.
With no channel compressor, it can achieve the same performance, suggesting the capability of the channel compressor in speeding up the kernel prediction without harming the performance.
Based on the above results, we set $C_m$ to 64 by default as a trade-off between
performance and efficiency.

For the downsampling version of \algname, we adopt different settings of $C_m$ in a ResNet-50 for image classification.
As shown in Table~\ref{tab:mid-cls}, with different $C_m$, the classification accuracy is relatively stable.
We choose $C_m=16$ that achieves the best performance and involves less computational cost.

We then investigate the influence of $k_{encoder}$ and $k_{reassembly}$.
Intuitively, increasing $k_{reassembly}$ also requires a larger $k_{encoder}$,
since the content encoder needs a large receptive field to predict a large
reassembly kernel.
As illustrated in Table~\ref{tab:k study}, increasing $k_{encoder}$ and $k_{reassembly}$
at the same time can boost the performance, while just enlarging one of them will not.
We summarize an empirical formula that $k_{encoder}=k_{reassembly}-2$, which is a good
choice in all the settings.
Though adopting a larger kernel size is shown helpful, we set $k_{reassembly}=5$ and
$k_{encoder}=3$ by default as a trade-off between performance and efficiency.

Other than the softmax function, we also test other alternatives in the kernel
normalizer, such as sigmoid or sigmoid with normalization.
As shown in Table~\ref{tab:softmax}, `Softmax' and `Sigmoid Normalized'
have the same performance and better than `Sigmoid', which shows that it is
crucial to normalize the reassembly kernel to be summed to 1.

\noindent
\textbf{How \algname~Works.}
We conduct a further qualitative study to figure out how \algname~works for upsampling and downsampling.
With trained Mask R-CNN models that adopts \algname~as the upsampling and downsampling operator, respectively,
we visualize the reassembling process in Figure~\ref{fig:arrow}.
Specifically, we sample some pixels in the feature map that the upsampling/downsampling process attained, and see which neighbors it is reassembled from.
The green circle denotes example locations and red dots indicate highly weighted sources during the reassembly.
For upsampling (see Figure~\ref{fig:arrow} (a)), in the FPN structure, the low-resolution feature map will be consecutively
upsampled for several times to a higher resolution, so a pixel in the upsampled
feature map reassembles information from a more larger region.
For downsampling (see Figure~\ref{fig:arrow} (b)), a input high-resolution feature map is downsampled for several times by \algname. In the process, pixels in a large region of the high-resolution
feature map is reassembled to attain a pixel on the low-resolution feature map. As a result, the receptive field of a downsampled feature map increases.
From the figure, we can clearly learn that \algname~is content-aware.
It tends to reassemble points with similar semantic information.
A location at the foreground instance, \eg, human body and wing of an airplane, prefers other points from the same instance, rather than other objects or nearby background.
For locations in the background regions which has weaker semantics, the reassembly is more uniform or just biased on points with similar low-level texture features.

% !TEX root = ../main.tex
% \vspace{-5pt}
\section{Conclusion}
\label{sec:conclusion}

We have presented Unified Content-Aware ReAssembly of FEatures (\algname), a universal, lightweight and highly effective feature reassembly operator.
It consistently boosts the performances on standard benchmarks in object detection, instance/semantic segmentation and inpainting by 2.5\% $AP_{box}$, 2.1\% $AP_{mask}$, 1.94\% mIoU, 1.35 dB, respectively.
More importantly, \algname~introduces little computational overhead and can be readily integrated into modern network architectures. It shows great potential to serve as a strong building block for future research.
Moreover, the current version of \algname~supports feature upsampling/downsampling by an integer factor. In our future work, \algname~will support feature reassembly with an arbitrary scale factor.
Then it could be more widely integrated for various network architectures.
\vspace{-0pt}

\noindent\textbf{Acknowledgements.}
This work is partially supported by the Collaborative Research Grant from SenseTime Group (CUHK Agreement No. TS1610626 \& No. TS1712093), the General Research Fund (GRF) of Hong Kong (No. 14236516 \& No. 14203518), Singapore MOE AcRF Tier 1 (M4012082.020), NTU SUG, and NTU NAP.

{\small
  \bibliographystyle{ieee_fullname}
  \bibliography{sections/egbib}
}

\newpage
\vspace{-30pt}
\begin{IEEEbiography}[{\includegraphics[width=1in,height=1.25in,clip,keepaspectratio]{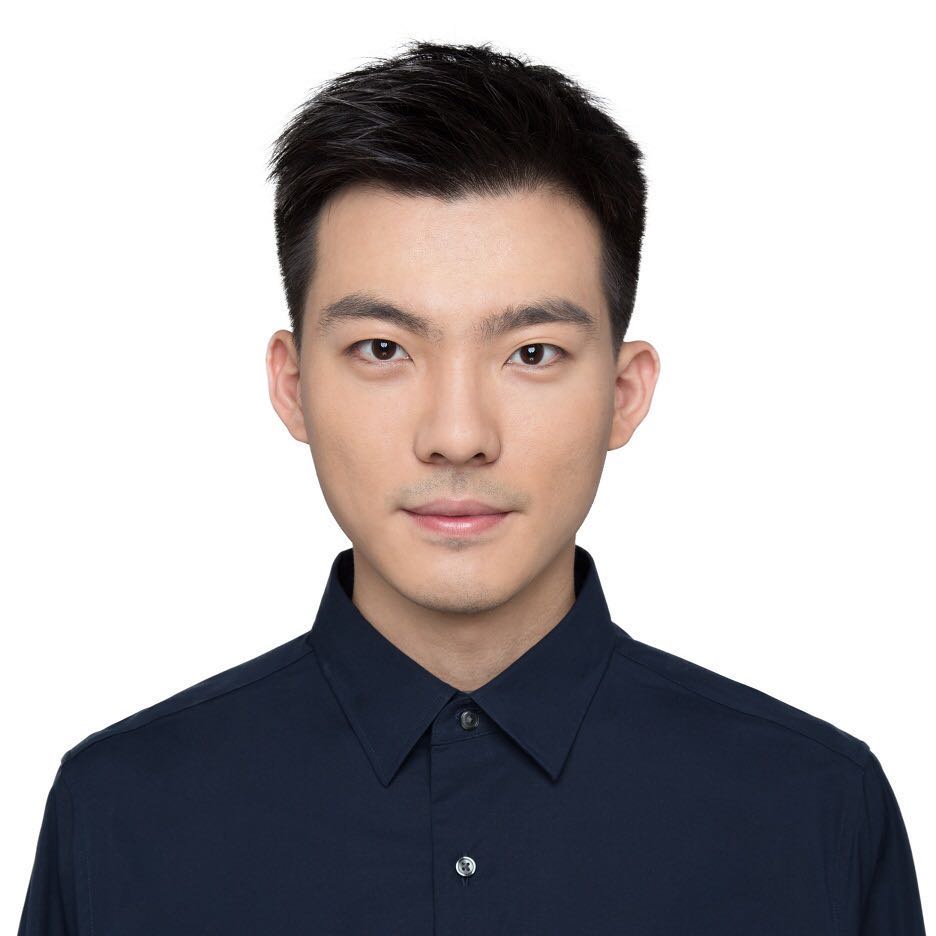}}]{Jiaqi Wang}
  is a fourth-year Ph.D. candidate in Multimedia Laboratory (MMLab) at the Chinese University of Hong Kong. He is supervised by Prof.
  Dahua Lin and works closely with Prof. Chen Change Loy. He received the B.Eng. degree from Sun Yat-Sen University in 2017 and began to pursue
  his Ph.D. degree funded by Hong Kong PhD Fellowship Scheme (HKPFS) in the same year. His research interests focus on object detection and
  instance segmentation.
\end{IEEEbiography}
\vspace{-30pt}
\begin{IEEEbiography}[{\includegraphics[width=1in,height=1.25in,clip,keepaspectratio]{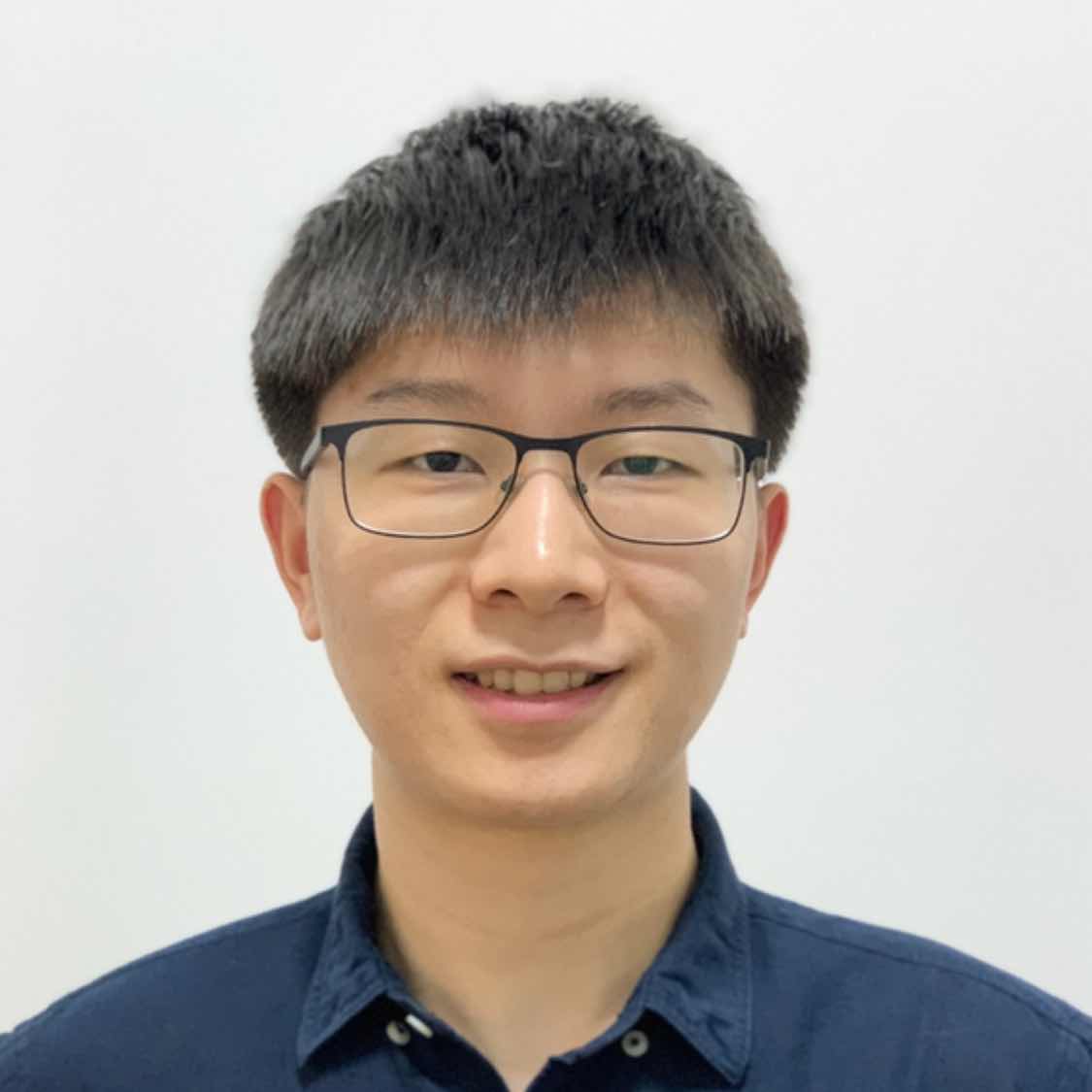}}]{Kai Chen}
  is a vice director at SenseTime, leading the algorithm platform team of EIG Research.
  He received the PhD degree in The Chinese University of Hong Kong in 2019, under the supervision of Prof. Dahua Lin and Chen Change Loy at MMLab. Before that, he received the B.Eng. degree from Tsinghua University in 2015.
  His research interests include computer vision and deep learning, particularly focusing on object detection, segmentation and video understanding.
\end{IEEEbiography}
%\newpage
\vspace{-30pt}
\begin{IEEEbiography}[{\includegraphics[width=1in,height=1.25in,clip,keepaspectratio]{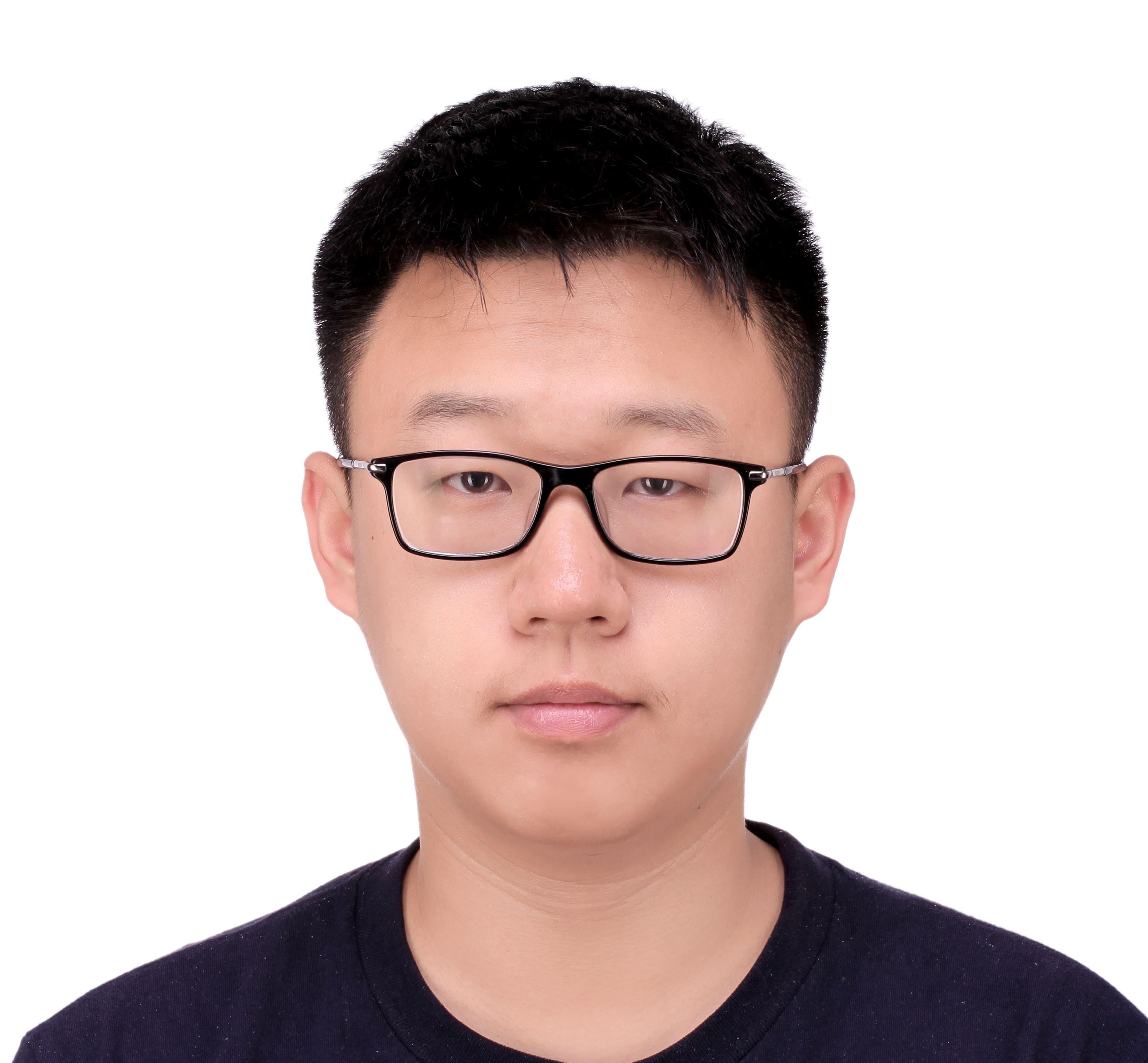}}]{Rui Xu}
  received the BEng degree in electronic engineering from Tsinghua University, in 2018. He is working toward the PhD degree in Multimedia Laboratory (MMLab) at the Chinese University of Hong Kong (CUHK). His research interests include computer vision and deep learning, especially for several topics in low-level vision like inpainting and image generation.
\end{IEEEbiography}
\vspace{-30pt}

\begin{IEEEbiography}[{\includegraphics[width=1in,height=1.25in,clip,keepaspectratio]{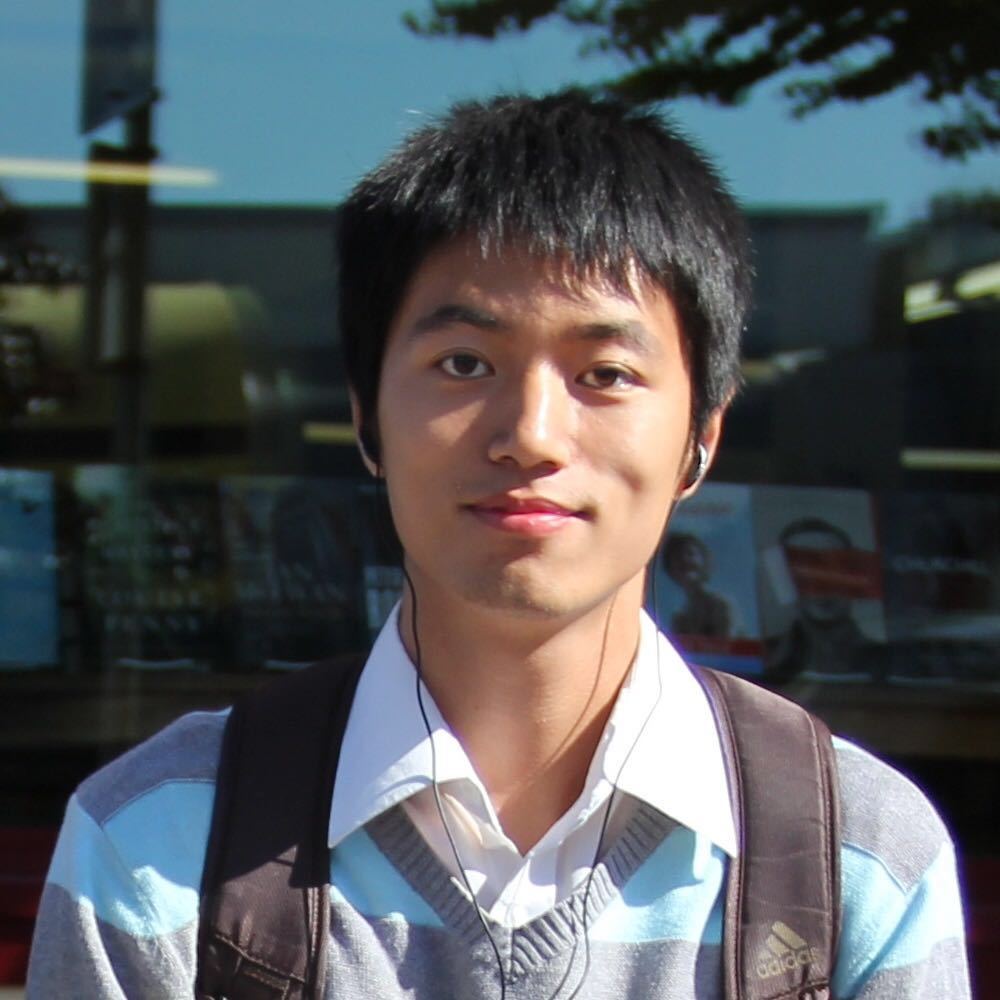}}]{Ziwei Liu}
  Ziwei Liu is currently a Nanyang Assistant Professor at Nanyang Technological University (NTU). Previously, he was a senior research fellow at the Chinese University of Hong Kong. Before that, Ziwei was a postdoctoral researcher at University of California, Berkeley, working with Prof. Stella Yu. Ziwei received his PhD from the Chinese University of Hong Kong in 2017, under the supervision of Prof. Xiaoou Tang and Prof. Xiaogang Wang. He has published over 50 papers (with more than 7,000 citations) on top-tier conferences and journals in relevant fields, including CVPR, ICCV, ECCV, AAAI, IROS, SIGGRAPH, T-PAMI, and TOG. He is the recipient of Microsoft Young Fellowship, Hong Kong PhD Fellowship, ICCV Young Researcher Award, and HKSTP best paper award.
\end{IEEEbiography}
\vspace{-30pt}

\begin{IEEEbiography}[{\includegraphics[width=1in,height=1.25in,clip,keepaspectratio]{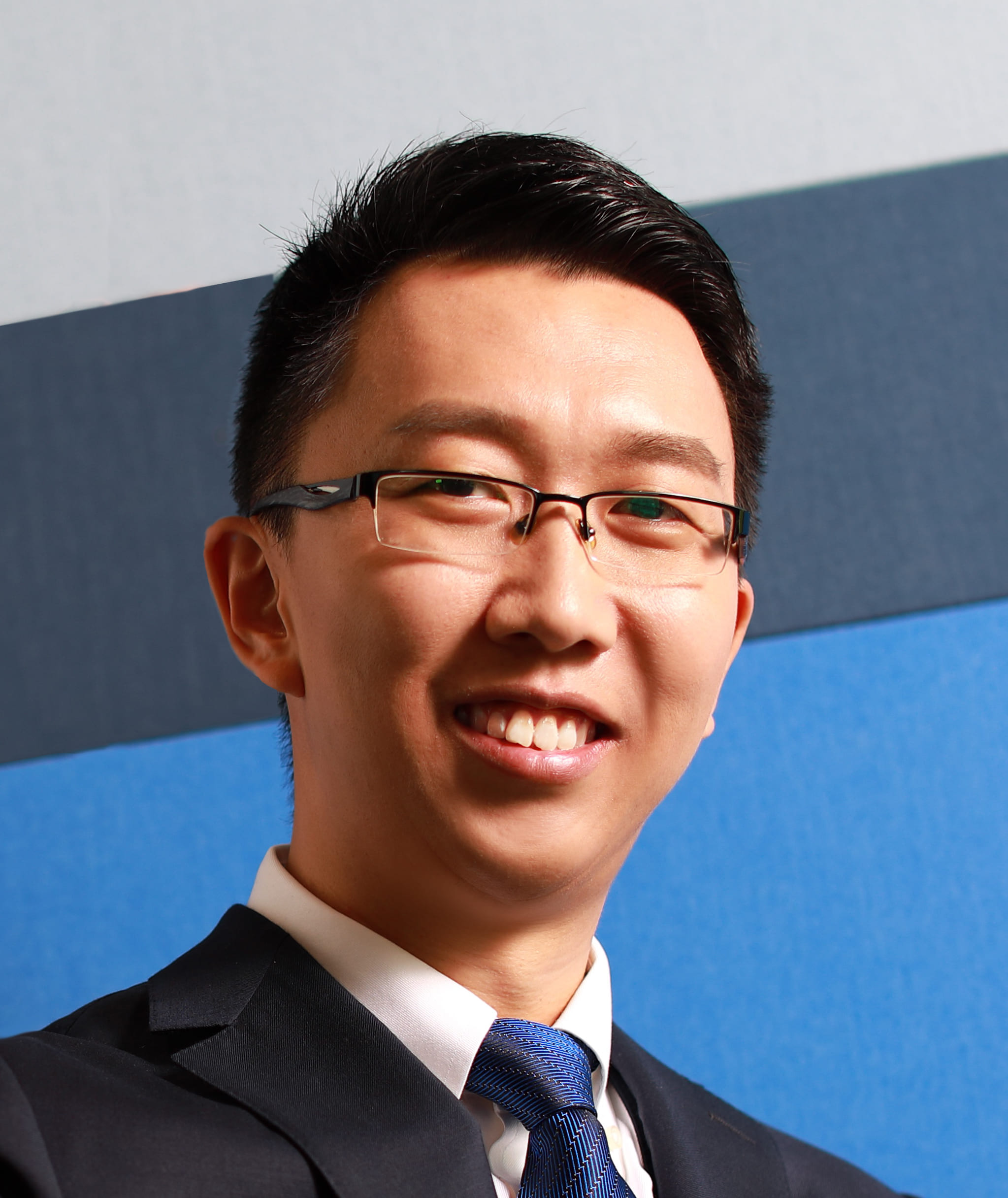}}]{Chen Change Loy} (Senior Member, IEEE) received the PhD degree in computer science from the Queen Mary University of London, in 2010. He is an associate professor with the School of Computer Science and Engineering, Nanyang Technological University. Prior to joining NTU, he served as a research assistant professor with the Department of Information Engineering, The Chinese University of Hong Kong, from 2013 to 2018. His research interests include computer vision and deep learning. He serves as an associate editor of the IEEE Transactions on Pattern Analysis and Machine Intelligence and the International Journal of Computer Vision. He also serves/served as an Area Chair of ICCV 2021, CVPR 2021, CVPR 2019, ECCV 2018, AAAI 2021 and BMVC 2018-2020.
\end{IEEEbiography}

\begin{IEEEbiography}[{\includegraphics[width=1in,height=1.25in,clip,keepaspectratio]{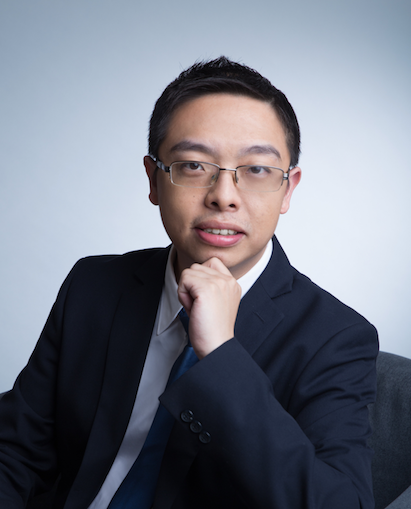}}]{Dahua Lin}
  Dahua Lin is an Associate Professor at the department of Information Engineering, the Chinese University of Hong Kong, and the Director of CUHK-SenseTime Joint Laboratory. He received the B.Eng. degree from the University of Science and Technology of China (USTC) in 2004, the M. Phil. degree from the Chinese University of Hong Kong (CUHK) in 2006, and the Ph.D. degree from Massachusetts Institute of Technology (MIT) in 2012. Prior to joining CUHK, he served as a Research Assistant Professor at Toyota Technological Institute at Chicago, from 2012 to 2014.
  His research interest covers computer vision and machine learning. He serves on the editorial board of the International Journel of Computer Vision (IJCV). He also serves as an area chair for multiple conferences, including ECCV 2018, ACM Multimedia 2018, BMVC 2018, CVPR 2019, BMVC 2019, AAAI 2020, and CVPR 2021.
\end{IEEEbiography}

%\vfill


\end{document}